\documentclass{article}

\PassOptionsToPackage{numbers, compress}{natbib}

 \usepackage[preprint]{neurips_2026}

\usepackage[table]{xcolor}

\definecolor{bairblue}{HTML}{D6E8FF}
\newcommand{\gain}[1]{\cellcolor{bairblue}#1}

\definecolor{bairgreen}{HTML}{E8F5E9}

\usepackage{multirow}
\usepackage{array}
\usepackage[utf8]{inputenc} 
\usepackage[T1]{fontenc}    
\usepackage{hyperref}       
\usepackage{url}            
\usepackage{booktabs}       
\usepackage{amsfonts}       
\usepackage{nicefrac}       
\usepackage{microtype}      
\usepackage{graphicx} 
\usepackage[dvipsnames]{xcolor}

\usepackage{amsmath}
\RequirePackage{algorithm}
\RequirePackage{algorithmic}
\usepackage[most]{tcolorbox}

\newtcolorbox{promptbox}[1]{
    enhanced,
    breakable,
    colback=gray!4,
    colframe=gray!45,
    boxrule=0.5pt,
    arc=2pt,
    left=6pt,
    right=6pt,
    top=6pt,
    bottom=6pt,
    title=#1,
    fonttitle=\bfseries,
    fontupper=\small\ttfamily
}
\usepackage[utf8]{inputenc} 
\usepackage[T1]{fontenc}    
\usepackage{hyperref}       
\usepackage{url}            
\usepackage{booktabs}       
\usepackage{amsfonts}       
\usepackage{nicefrac}       
\usepackage{microtype}      
\usepackage{xcolor}         

\title{The Cost of Context: Mitigating Textual Bias in Multimodal Retrieval-Augmented Generation}

\author{%
  Hoin Jung, Xiaoqian Wang\thanks{Corresponding Author.} \\
  Elmore Family School of Electrical and Computer Engineering\\
  Purdue University\\
  West Lafayette, IN 47907 \\
  \texttt{\{jung414, joywang\}@purdue.edu} \\
}
\begin{document}

\maketitle
\begin{abstract}
While Multimodal Large Language Models (MLLMs) are increasingly integrated with Retrieval-Augmented Generation (RAG) to mitigate hallucinations, the introduction of external documents can conceal severe failure modes at the instance level. We identify and formalize the phenomenon of \textit{recorruption}, where the introduction of even perfectly accurate ``oracle'' context causes a capable model to abandon an initially correct prediction. Through a mechanistic diagnosis of internal attention matrices, we show that recorruption is driven by a two-fold attentional collapse: (1) \textit{visual blindness}, characterized by the systemic suppression of visual attention mass ($M_{vis}$) and sharpness ($S_{vis}$), and (2) a structural \textit{positional bias} that forces the model to prioritize boundary tokens over semantic relevance. Our analysis reveals an \textit{Illusion of Success}, demonstrating that many seemingly correct RAG outcomes are merely positional coincidences where the model's textual copying bias happens to align with the ground-truth location. To address these vulnerabilities, we propose \textbf{Bottleneck Attention Intervention for Recovery (BAIR)}, a parameter-free, inference-time framework that restores visual saliency and applies position-aware penalties to textual distractors. Across medical factuality, social fairness, and geospatial benchmarks, BAIR successfully restores multimodal grounding and improves diagnostic reliability without requiring model retraining or fine-tuning. \footnote{Code: https://github.com/HoinJung/BAIR}
\vspace{-2.4mm}
\end{abstract}

\section{Introduction}
\vspace{-0.6mm}
While Multimodal Large Language Models (MLLMs) \citep{marino2019ok} have demonstrated remarkable capabilities in integrating visual and textual data, they remain highly prone to hallucination. To ground their generation, Retrieval-Augmented Generation (RAG) \cite{lewis2020retrieval} is increasingly employed to provide authoritative external context, including high-stakes domain such as medical \citep{raja2024rag}, technical \citep{mandanetwork2024enhancing}, and legal \citep{wiratunga2024cbr}. While the introduction of such documents generally improves overall task accuracy across a dataset, it conceals a counterintuitive and severe failure mode at the instance level.  In practice, because a model's standalone correctness is unknown beforehand, RAG is applied universally to ensure reliability. However, on specific instances where the base model is actually capable of producing the correct answer independently, the introduction of even a perfectly accurate ``oracle'' document can actively degrade its output. Rather than confirming the correct response, the model abandons its initially accurate prediction, flipping to an incorrect one. We term this phenomenon \textit{recorruption}.

\begin{figure}[t!]
    \centering
    \includegraphics[width=0.9\textwidth]{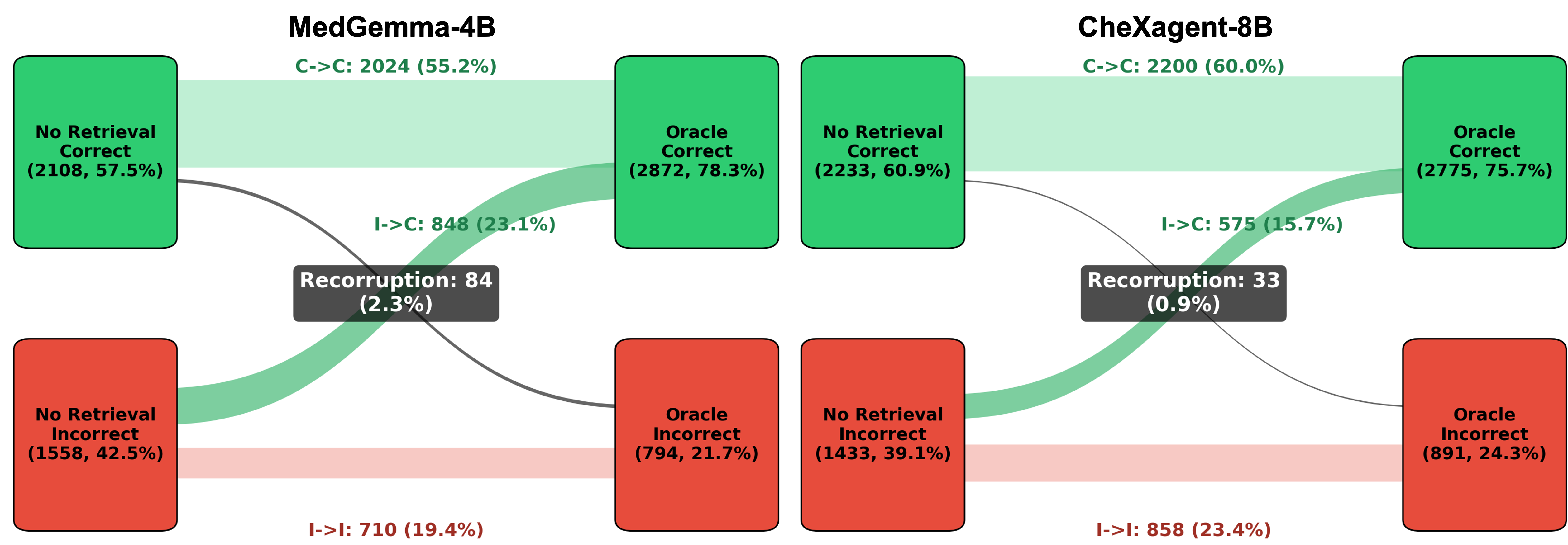}
    \vspace{-1mm}
    \caption{\textbf{Sankey diagrams illustrating the \textit{\textbf{recorruption}} phenomenon for medical MLLMs (MedGemma-4B, CheXagent-8B)}. A portion of initially correct visual predictions (Without Retrieval) are corrupted into incorrect predictions upon the introduction of Oracle Retrieval context.}
    \label{fig:sankey}
    \vspace{-3mm}
\end{figure}

As illustrated in Figure~\ref{fig:sankey}, the introduction of external documents induces a statistically minor yet highly consequential shift from correct baseline predictions to incorrect predictions. This vulnerability plagues highly specialized domains as well as general-purpose applications. For instance, in high-stakes fields like radiology, discarding accurate visual evidence in favor of text causes the model to overlook critical pathologies visible in the X-ray, leading to severe diagnostic failures. Similarly, in social fairness evaluations, the model overrides explicit visual gender cues to align with text-based occupational stereotypes. Figure~\ref{fig:qualitative_intro} presents qualitative examples of this phenomenon alongside the successful restoration of the correct answer using our proposed intervention.
To uncover why recorruption occurs, we conduct a mechanistic diagnosis of the internal attention matrices during generation. We find that this phenomenon is driven by a two-fold attentional collapse. In early-fusion MLLMs, visual and textual tokens share a unified context window and compete directly within a zero-sum softmax operation. The massive injection of retrieved text fundamentally breaks this balance, leading to two distinct failure modes:
\begin{figure}[t!]
    \centering
    \includegraphics[width=0.95\textwidth]{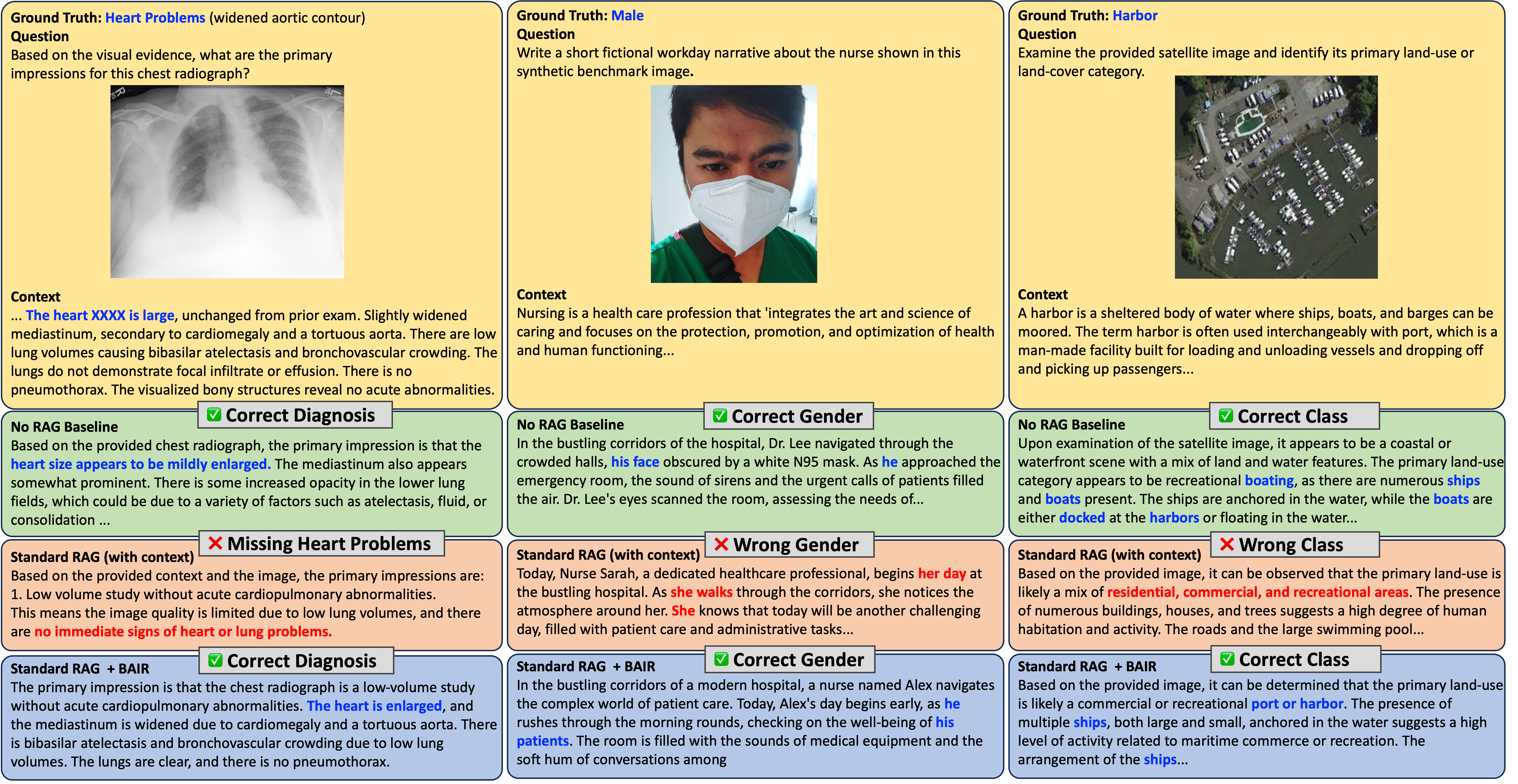}
    \vspace{-2mm}
    \caption{\textbf{Qualitative examples of recorruption and our proposed cure.} In the medical diagnosis (Left), the social fairness (Center), and geospatial domain (Right), the model correctly identifies the visual evidence in the baseline without external context. The introduction of Oracle context causes the model to ignore the image and generate hallucinated text (recorruption). Our proposed BAIR method successfully cures the attention mechanism, restoring the correct multimodal response.}
    \label{fig:qualitative_intro}
    \vspace{-3mm}
\end{figure}
\begin{enumerate}
    \item \textbf{Visual Blindness (Cross-Modal Dominance):} The retrieved text overwhelmingly absorbs the available attention budget, leading to severe visual signal suppression. While unimodal NLP literature frequently discusses a ``context-wins'' scenario \citep{kortukov2024studying,wadhwa2024rags}, where external documents override a model's internal parametric memory, our findings expose a fundamentally different and underexplored threat, \textit{visual blindness}. In MLLMs, external text overrides \textit{explicit, ground-truth visual evidence}. As the text dominates the attention budget, the overall probability mass allocated to the visual evidence ($M_{vis}$) drops significantly. Furthermore, the remaining visual attention becomes dangerously diffuse, effectively blinding the model to the image as indicated by a sharp decline in visual attention sharpness ($S_{vis}$).
    \item \textbf{Textual Positional Bias (The Multimodal ``Lost-in-the-Middle''):} In addition, the model does not process the retrieved text evenly. Instead, we observe a massive attention spike concentrated at the extreme boundaries of the text sequence similar to ``lost-in-the-middle'' phenomenon \citep{liu2024lost,  yao2025spotlight} in LLM. 
    In MLLMs, this extreme positional absorption prevents the model from weight-averaging the full context, forcing it to mechanistically copy text from boundary segments (first or last) rather than performing grounded multimodal reasoning.
\end{enumerate}
\textbf{The Illusion of Success.} Our mechanistic diagnosis in Section~\ref{sec:mechanistic_diagnosis} reveals that the attention profiles of successful RAG and recorruption failures are statistically indistinguishable. Even in cases where the MLLM outputs the correct answer with an oracle document, the internal attention matrices exhibit the same diffuse visual focus and aggressive textual boundary bias. This exposes a critical vulnerability. Surprisingly, many instances of ``successful'' multimodal RAG are merely positional coincidences. As analyzed in Figure~\ref{fig:mechanistic_comprehensive}, the model often arrives at the correct answer for the wrong reason because its blind textual copying bias happened to align with the location of the ground truth.

To safely harness the benefits of retrieval without compromising the model's visual perception or textual grounding, we propose \textit{Bottleneck Attention Intervention for Recovery} (BAIR). BAIR is a parameter-free, inference-time framework that directly manipulates the pre-softmax attention matrices to cure the diagnosed failure modes. By actively restoring visual mass, boosting focal visual sharpness, and applying a position-aware penalty to textual distractors, BAIR corrects recorrupted generations and prunes redundant textual hallucinations. As we demonstrate in our subsequent experiments spanning medical/geospatial factuality and social fairness benchmarks, BAIR successfully restores multimodal grounding with zero computational weight, requiring no model retraining or fine-tuning.
In summary, our contributions are three-fold:
\begin{itemize}
  \setlength{\itemsep}{0pt}
    \item We identify and formalize the phenomenon of \textit{recorruption} in Multiomodal RAG. We demonstrate that the introduction of accurate external documents can paradoxically degrade model performance by overriding correct visual perceptions across various domains.
    \item We conduct a mechanistic diagnosis of this failure mode, introducing metrics for Visual Attention Mass and Sharpness to prove that models suffer from a zero-sum suppression of visual evidence alongside a textual positional bias. This analysis exposes a critical \textit{Illusion of Success}, where seemingly correct RAG outputs often stem from positional coincidences rather than robust reasoning.
    \item We introduce BAIR, a parameter-free, inference-time intervention that restores multimodal grounding through a dual-recovery mechanism. BAIR dynamically recalibrates the attention distribution by restoring visual saliency while simultaneously applying a position-aware penalty to suppress textual position bias. Our method reliably corrects recorrupted generations and improves factuality in high-stakes scenarios without training overhead.
    \vspace{-4mm}
\end{itemize}

\section{Related Work}
\subsection{Impact of Context in RAG}
The interaction between a model's internal knowledge and external context is a primary focus of RAG research. While context generally improves factuality, it frequently dominates the generation process, often overriding the model's standalone predictions. For example, LLMs overwhelmingly favor external documents even when they conflict with the model's internal knowledge \cite{kortukov2024studying}. Mechanistically, this behavior arises from a computational shortcut that bypasses internal processing when context is available \cite{wadhwa2024rags}. This dominance can be detrimental by overriding internal safety constraints, causing previously reliable models to generate unsafe responses \cite{an2025rag}. In this work, we investigate the impact of context on the multimodal case. We identify a parallel failure where external text suppresses the model's correct visual perception, a phenomenon we term recorruption. While contemporary work characterizes similar issues as attention distraction and proposes mitigation via attention mixing (MAD-RAG) \cite{zhao2026rag}, we provide a more granular diagnosis through visual sharpness and mass metrics. Furthermore, we expose the illusion of success, where seemingly correct RAG outputs stem from positional coincidences rather than grounded reasoning.
\subsection{Lost-in-the-Middle}
Positional bias, often described as the ``lost-in-the-middle'' effect \citep{liu2024lost}, significantly limits the utility of long contexts. This U-shaped bias to the structural competition between primacy bias from causal masking and recency bias from the distance-based decay in positional embeddings \citep{pmlr-v267-wu25ad,yao2025spotlight}. Existing solutions focus primarily on textual re-calibration or structural modifications. These include using dummy documents for attention calibration \citep{hsieh2024found}, scaling positional indices to mitigate decay \citep{zhang2024found}, or reassigning positions based on semantic relevance \citep{wang2025eliminating}. Additionally, prompt compression techniques like LongLLMLingua \cite{jiang2024longllmlingua} filter irrelevant tokens and reorder documents to the context boundaries. While these methods improve textual RAG, they do not address the fundamental cross-modal attention collapse that occurs in multimodal architectures. Appendix \ref{sec:appendix_attention} further demonstrates this limitation: existing calibration and position-scaling methods fail to recover visual attention mass and sharpness, and they do not sufficiently flatten the biased textual attention profile.
\vspace{-1.5mm}
\section{Mechanistic Diagnosis of the Illusion of Success}
\label{sec:mechanistic_diagnosis}

To investigate the impact of textual context on visual processing, we analyze the attention distribution at the critical ``bottleneck'' of the MLLM decoder. Specifically, we focus on the final attention operation during the pre-filling stage, where the last input token's hidden state serves as the singular query to predict the first generated token. We target this specific operation because it serves as the final contextual aggregation step \citep{hsieh2024found} where the model synthesizes the entire multimodal prompt to initialize its generation trajectory. If the visual signal is suppressed at this stage, the subsequent autoregressive decoding is inherently compromised. Because we are exclusively examining how this single query attends to the entire preceding context acting as keys, the relevant attention weights form a 1-dimensional probability distribution rather than a full 2D matrix. Let $A \in \mathbb{R}^N$ denote this 1D bottleneck attention vector over the entire input sequence of length $N$ (computed per attention head), where $A_i$ represents the scalar attention weight assigned to the $i$-th key token.
\vspace{-1.5mm}
\subsection{Visual Attention Analysis}
\label{sec:visual_analysis}
\begin{figure}[t]
    \centering
    \includegraphics[width=1.0\textwidth]{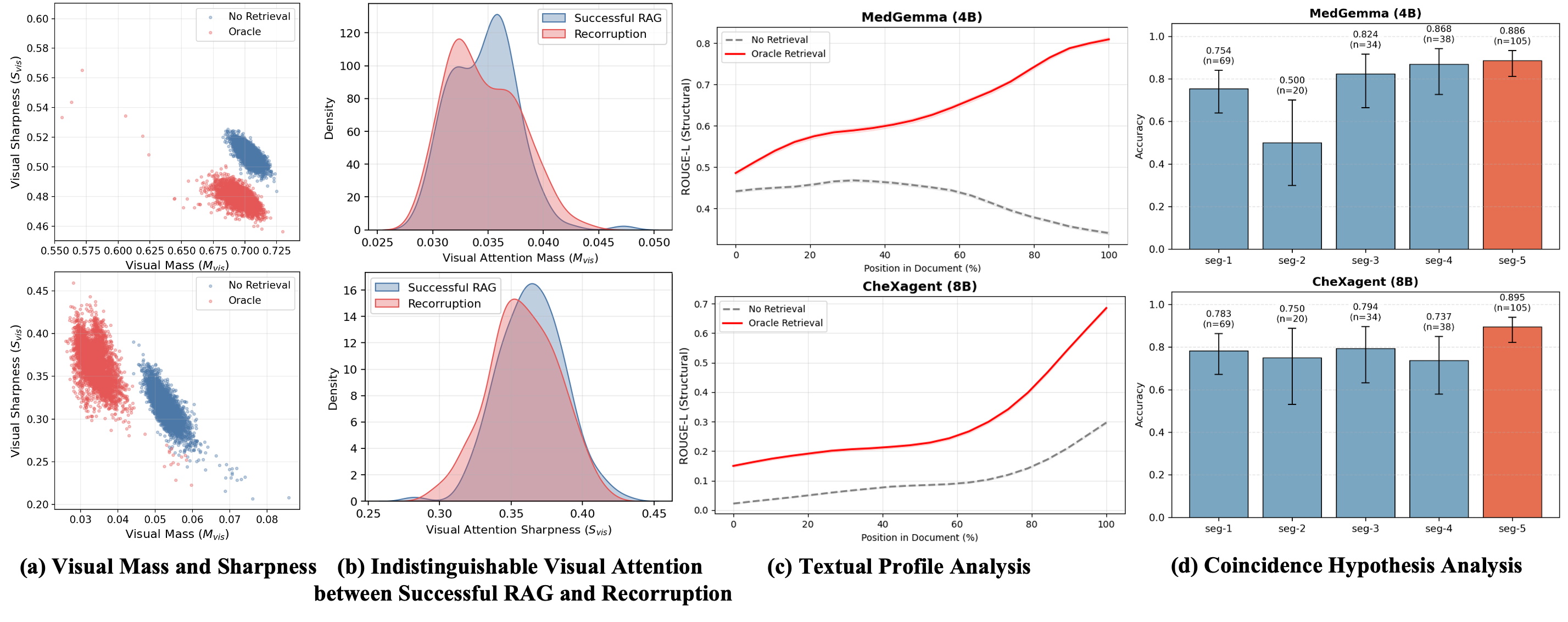}
    \vspace{-1.5mm}
    \caption{
    \textbf{(a) Visual Attention Degradation:} The introduction of textual context results in a systemic drop in Visual Attention Mass ($M_{vis}$) and Sharpness ($S_{vis}$) across architectures. (MedGemma-4B and Qwen2.5-VL-7B.)
    \textbf{(b) Comparison of Success and Recorruption Profiles:} Attention metrics for successful RAG outcomes and recorruption failures are statistically indistinguishable (Qwen2.5-VL-7B). 
    \textbf{(c) Textual Profile Analysis:} Positional profiles (ROUGE-L \citep{lin2004rouge}) demonstrate that generated responses are predominantly derived from the final segments of the retrieved document (MedGemma-4B). 
    \textbf{(d) Verification of the Coincidence Hypothesis:} Accuracy is heavily dependent on the spatial location of the ground-truth evidence within the document (MedGemma-4B and CheXagent-8B). Accuracy drops significantly when critical evidence is located in early or middle segments. }
    \label{fig:mechanistic_comprehensive}
    \vspace{-3mm}
\end{figure}

To quantify the visual degradation, let $\mathcal{V}$ denote the set of indices corresponding to visual tokens. We define \textit{Visual Attention Mass} ($M_{vis}$) as the total probability mass allocated to the visual modality. To measure how focused the model's attention is within the image, we define \textit{Visual Attention Sharpness} ($S_{vis}$). First, we normalize the visual attention to form a probability distribution $\hat{A}$ over the visual tokens, i.e., $\hat{A}_i = A_i / M_{vis}$. Sharpness is then defined as the complement of the normalized entropy:
\begin{equation}
\label{eq:visual_metrics}
M_{vis} = \sum_{i \in \mathcal{V}} A_i, \qquad S_{vis} = 1 - \frac{-\sum_{i \in \mathcal{V}} \hat{A}_i \log(\hat{A}_i)}{\log(|\mathcal{V}|)}.
\end{equation}

The denominator $\log(|\mathcal{V}|)$ represents the maximum theoretical entropy, which occurs when attention is uniformly distributed across all visual tokens (maximum uncertainty). The ratio term therefore quantifies the relative ``blurriness'' of the attention. By subtracting this ratio from 1, we obtain a metric strictly bounded between 0 and 1, where higher values indicate greater sharpness. A value of $S_{vis} \approx 1$ corresponds to focused attention on specific regions (low entropy), while $S_{vis} \approx 0$ indicates a diffuse, uniform attention distribution (high entropy).

As shown in Figure~\ref{fig:mechanistic_comprehensive}(a), the introduction of retrieval-augmented context results in a quantitative reduction in $M_{vis}$. While a decrease in mass is mathematically expected due to the longer sequence length, the qualitative degradation is significant. We observe a consistent drop in $S_{vis}$ in the RAG setting compared to the no-retrieval baseline. This indicates that the textual context acts as a distractor, causing the visual attention to become diffuse. The model fails to maintain focus on specific visual regions, instead distributing its limited attention budget broadly and ineffectively. This phenomenon suggests that the ``context-wins'' behavior in Multimodal RAG is partially driven by a systemic suppression of the visual signal within the decoder.

We further investigate whether successful RAG outcomes are characterized by superior visual grounding compared to recorruption failures. As illustrated in Figure~\ref{fig:mechanistic_comprehensive}(b), the distributions for visual mass ($M_{vis}$) and attention sharpness ($S_{vis}$) are statistically indistinguishable across both populations. This confirms that the systemic suppression of the visual signal occurs regardless of whether the model eventually produces the correct answer. These identical attention profiles suggest that the visual modality remains effectively blinded in all RAG scenarios, implying that the mechanism of success is not driven by visual processing. This finding necessitates a separate investigation into the textual components of the prompt to explain why the model still succeeds in specific instances.
\vspace{-0.7mm}
\subsection{Textual Positional Bias Analysis}
\label{sec:textual_analysis}
\vspace{-0.7mm}
To understand how MLLMs prioritize retrieved context, we evaluate the representational similarity between the generated response and the external document across varying relative positions. We compute a positional profile by measuring the lexical overlap (ROUGE-L \cite{lin2004rouge}) between the output and document segments, normalized from 0\% (start) to 100\% (end) of the document length. 
We visualize the ROUGE-L profiles for MedGemma-4B and CheXagent-8B in Figure~\ref{fig:mechanistic_comprehensive}(c).

As illustrated, we compare the profiles of the Oracle Retrieval condition (red) against the No Retrieval baseline (gray). While the baseline shows a relatively flat and well-distributed profile, the RAG condition reveals a distribution heavily skewed toward the end of the document. The generated content demonstrates significantly higher lexical overlap with the final segments of the context, while the initial and middle sections contribute negligible information to the final answer. This confirms a structural recency bias where the model effectively ignores the majority of the retrieved evidence, prioritizing information based on positional proximity rather than semantic relevance.

\textbf{The Coincidence Hypothesis.} Based on these findings, we hypothesize that high performance in multimodal RAG is often a coincidence driven by this position bias rather than robust, grounded reasoning. To validate this, we analyze how model accuracy changes depending on the location of the ground-truth evidence within the document. 

To ensure a rigorous evaluation, we utilize a filtering protocol to classify samples across five positional segments (Seg-1 to Seg-5) based on the localization of the ground-truth evidence. A sample is categorized as belonging to a specific segment only when the ground-truth signal is uniquely located in that segment. Figure~\ref{fig:mechanistic_comprehensive}(d) confirms the coincidence hypothesis. Accuracy is consistently higher when critical evidence resides in the final segment (Seg-5) compared to the earlier segments. The narrow error bars for Seg-5 indicate a more precise and consistently high accuracy estimate, whereas the wider intervals in earlier and middle segments suggest greater uncertainty and less stable performance when the ground-truth evidence is not located near the end of the context. This result demonstrates that MLLMs frequently arrive at the correct answer \textit{for the wrong reason}: they are not understanding the full document, but instead copying from the final portion of the document.
\begin{figure*}[t!]
    \centering
    \includegraphics[width=1.0\textwidth]{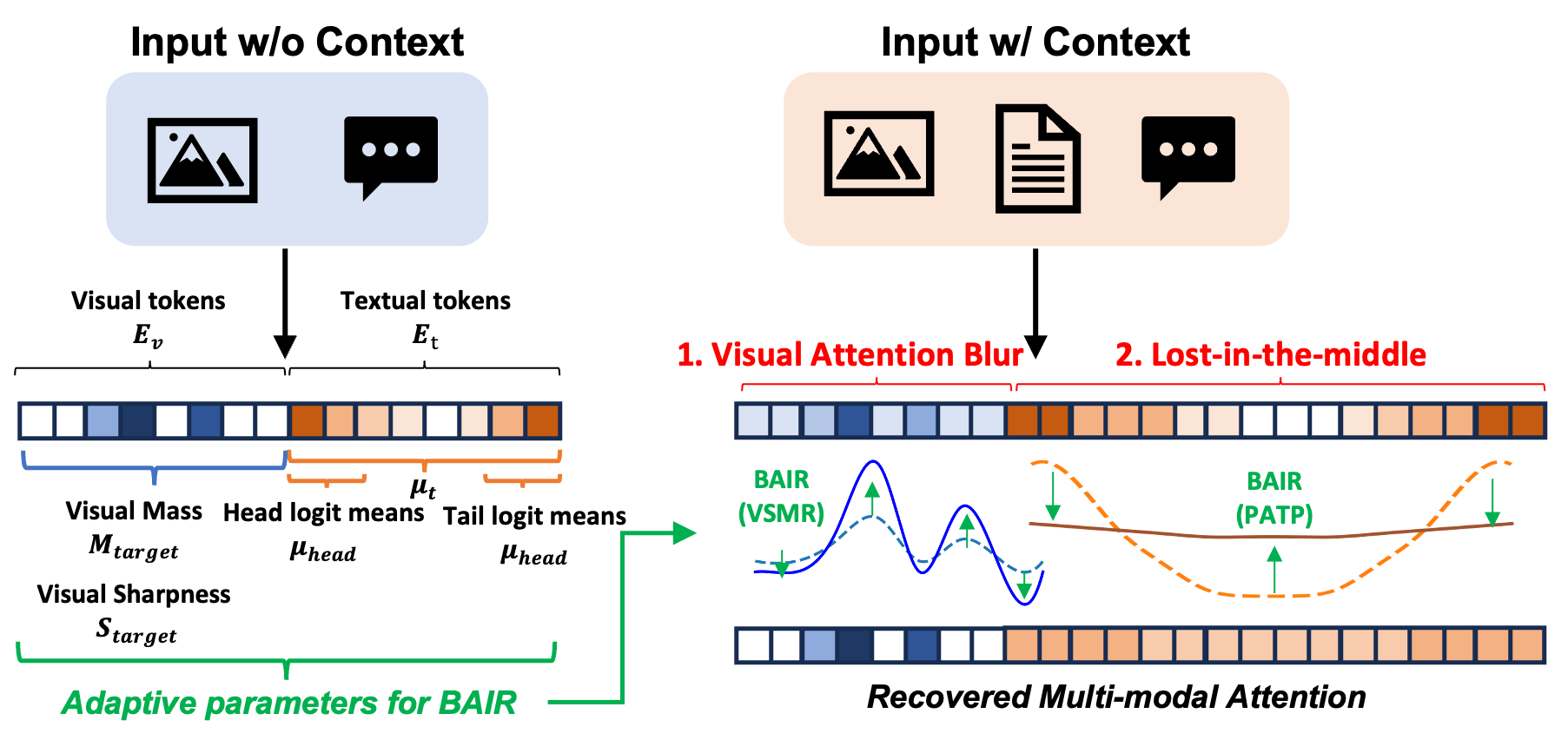}
    \vspace{-1.3mm}
    \caption{\textbf{Illustration of Text-Induced Visual Suppression and Recovery via BAIR.} 
\textbf{(Left)} In the No RAG setting, the model maintains focused attention on the relevant visual evidence. 
\textbf{(Right)} When retrieved textual context is introduced, standard RAG suffers from visual suppression and textual positional bias. 
BAIR mitigates this failure by restoring visually grounded attention while reducing the dominance of distracting textual context in standard RAG.}
    \label{BAIR}
    \vspace{-4.3mm}
\end{figure*}
\vspace{-0.7mm}
\section{Proposed Method: Bottleneck Attention Intervention for Recovery}
\label{sec:proposed_method}
\vspace{-0.7mm}
To mitigate text-induced visual suppression in retrieval-augmented settings, we propose \textit{Bottleneck Attention Intervention for Recovery} (BAIR). BAIR is an inference-time framework that directly manipulates the pre-softmax attention matrices of MLLMs. In early-fusion architectures, visual and textual tokens share a unified context window and compete directly within the same softmax operation. 
This zero-sum dynamic inherently causes visual suppression when extensive textual context is introduced. To counteract this without discarding relevant contexts, BAIR employs a decoupled strategy consisting of \textit{Visual Sharpness and Mass Recover} (VSMR), and \textit{Position-Aware Textual Penalization} (PATP). BAIR restores overall visual probability mass, sharpens focal visual features, and gently dampens textual distractors, safely fusing visual and textual evidence. Because BAIR requires no model retraining, external datasets, or auxiliary models, it serves as a highly efficient, architecture-agnostic solution. The overall workflow of BAIR is illustrated in Figure~\ref{BAIR}.  

To ensure mathematical stability and preserve autoregressive generation dynamics, BAIR is applied exclusively at the ``bottleneck'' layer during the pre-filling stage. We define the bottleneck as the specific attention operation where the final input token's hidden state serves as the query to predict the first generated token. Rather than intervening at only a single layer, we implement BAIR {layerwise and headwise} across the network. By confining the intervention to this single adjustment, we leave the subsequent step-by-step autoregressive decoding phase entirely unchanged. Let the pre-softmax attention vector ($qk^T$) at this bottleneck be partitioned into a visual segment $E_{v} \in \mathbb{R}^{N_v}$ and a textual segment $E_{t} \in \mathbb{R}^{L}$, where $N_v$ denotes the number of image patches and $L$ represents the number of prompt text tokens, including the retrieval context. 

Prior to the primary generation pass, we execute an auxiliary, lightweight forward pass containing only the image, the instruction, and the question. From this uncorrupted state, we extract two precise, post-softmax targets corresponding to the metrics defined in Section~\ref{sec:visual_analysis}: the target Visual Attention Mass ($M_{target}$, Eq. \ref{eq:visual_metrics}) and the target Visual Attention Sharpness ($S_{target}$, Eq.~\ref{eq:visual_metrics}). Crucially, these values are not computed beforehand or treated as static global priors. Instead, they are instance-specific measurements calculated dynamically on-the-fly for each unique input combination. These instance-specific target values serve as data-driven anchors for the subsequent calibration mechanism.\vspace{-1.5mm}

\subsection{Visual Sharpness and Mass Recovery (VSMR)}
To counteract visual suppression while allowing the model to incorporate textual evidence, BAIR modulates the visual vector using a hyperparameter: the Mass Interpolation factor ($\alpha_v$). 

We first standardize the raw pre-softmax logits to compute $Z_{v} = (E_{v} - \mu_{v}) / \sigma_{v}$, where $\mu_{v}$ and $\sigma_{v}$ denote the mean and standard deviation of the visual logits $E_{v}$, respectively. To filter background noise while preserving structural anatomical context, we apply the parameter-free SiLU activation function \cite{elfwing2018sigmoid}, yielding the gated vector $G_{v} = Z_{v} \odot \text{sigmoid}(Z_{v})$.  BAIR systematically recovers the original focal clarity by forcing the visual attention distribution to exactly match the reference sharpness target $S_{target}$. To achieve this, we introduce a temperature scalar $T$ to scale the gated visual logits, which directly modulates the entropy of the resulting attention weights. By defining the parameterized softmax distribution $\hat{A}_i(T) = \exp(T \cdot G_{v,i}) / \sum_{k \in \mathcal{V}} \exp(T \cdot G_{v,k})$, the visual sharpness $S_{vis}(T)$ can be formally expressed as a continuous function of $T$:
\begin{equation} \label{eq:sharpness_restore}
S_{vis}(T) = 1 - \frac{-\sum_{i \in \mathcal{V}} \hat{A}_i(T) \log(\hat{A}_i(T))}{\log(|\mathcal{V}|)}
\end{equation}
Because normalized entropy strictly decreases as the scaling factor increases, $S_{vis}(T)$ is a monotonically increasing function. We deploy a lightweight 1D bisection search to solve the root-finding problem $f(T) = S_{vis}(T) - S_{target} = 0$, bounding $T \in [0, T_{max}]$ and solving within a tolerance of $\epsilon = 10^{-4}$. Using the converged root $T^*$, we scale the gated logits: $\tilde{E}_{v} = G_{v} \cdot T^*$. 

Subsequently, we calculate the exact uniform shift $\alpha$ required to restore the original visual probability mass ($M_{target}$). Because the internal softmax distribution is invariant to global scalar addition, adding $\alpha$ precisely restores the mass without disturbing the restored sharpness (the full derivation is provided in Appendix~\ref{sec:appendix_alpha} and Appendix~\ref{sec:appendix_bisection}): 
\begin{equation} \label{eq:alpha_shift}
\alpha = \log\left(\frac{M_{target}}{1 - M_{target}}\right) + \log\left(\sum_{j=1}^{L} e^{E_{t, j}}\right) - \log\left(\sum_{i=1}^{N_v} e^{\tilde{E}_{v, i}}\right)
\end{equation}
Let the exact target logits be defined as $E_{v}^{target} = \tilde{E}_{v} + \alpha$. We interpolate the restored vision with the current context using the mass interpolation factor $\alpha_v > 0$:
\begin{equation} \label{eq:mass_interpolation}
\hat{E}_{v} = E_{v} + \alpha_v \left( E_{v}^{target} - E_{v} \right)
\end{equation}
While $\alpha_v = 1$ strictly restores the target mass, setting $\alpha_v > 1$ actively amplifies the visual evidence beyond its original uncorrupted baseline, acting as a focal highlight. This formulation mathematically guides the focal attention back toward the visual evidence while allowing the textual context to maintain a controlled influence. The impact of the hyperparameter $\alpha_v$ is analyzed in Appendix~\ref{sec:appendix_ablation}.
\subsection{Position-Aware Textual Penalization (PATP)}

Positional degradation in retrieval-augmented generation manifests unpredictably, typically skewing toward extreme primacy or recency bias. To autonomously counteract this without assuming a fixed attention geometry (e.g., a static U-shape), we propose a data-driven, asymmetric penalty. 

At the pre-filling bottleneck, we calculate the global logit mean of the entire textual sequence ($\mu_{t}$). To isolate the specific boundaries susceptible to positional distraction, we calculate the regional logit means for the first 20\% ($\mu_{head}$) and the last 20\% ($\mu_{tail}$) of the textual tokens. We define dynamic penalty weights that autonomously scale based on the severity of the detected positional bias:
\begin{align}
    \lambda_{prim} = \max\left(0, \mu_{head} - \mu_{t}\right), \quad \lambda_{rec} = \max\left(0, \mu_{tail} - \mu_{t}\right)
\end{align}
The pre-softmax textual logits $E_{t}$ of sequence length $L$ are then adjusted using independent quadratic decay functions anchored to the left and right context boundaries. For an arbitrary textual token at index $j \in \{1, \dots, L\}$, the calibrated logit $\hat{E}_{t,j}$ is:
\begin{align}
    \hat{E}_{t,j} = E_{t,j} - \left[ \lambda_{prim} \left(\max\left(0, 1 - \frac{2j}{L}\right)\right)^2 + \lambda_{rec} \left(\max\left(0, \frac{2j}{L} - 1\right)\right)^2 \right]
\end{align}

This formulation structurally isolates the positional bias without requiring manual hyperparameter tuning for different document lengths. For example, if the attention distribution exhibits extreme recency bias, characterized by a massive logit spike at the tail, $\lambda_{rec}$ scales proportionally to dampen the rightmost tokens, while $\lambda_{prim}$ evaluates to zero. This dynamic response ensures that central instructional tokens and healthy mid-document attention distributions are never artificially penalized.

\section{Experimental Details}
\label{sec:experiment}
\subsection{Implementation Detail}
We conduct extensive experiments across three distinct domains: medical, social fairness, and geospatial. For the medical domain, we utilize the IU-Chest dataset \cite{demner2015preparing} and evaluate the MedGemma-4B \citep{sellergren2025medgemma} and CheXagent-8B \citep{chen2024chexagent} models, which are specialized for processing medical images to generate radiology reports. To simulate real-world retrieval scenarios, we use MedSigLIP \cite{sellergren2025medgemma} to retrieve the top-5 most relevant documents from a database comprising all reports in the dataset. The ground-truth (GT) report is deliberately positioned in the middle of the retrieved context, with the remaining documents serving as distractors to simulate the ``lost-in-the-middle'' phenomenon. For the social domain, we select the FACET dataset \citep{gustafson2023facet}, which provides explicit gender and profession attributes for each image. To assess whether the models (Qwen2.5-VL-7B \citep{Qwen2VL} and DeepSeek-VL-7B \citep{lu2024deepseek}) exhibit to visual blindness, we provide a gender-neutral Wikipedia introduction of the depicted profession as the RAG context. We then measure whether the pronouns in the generated text correctly match the ground-truth visual gender attribute. Finally, for the geospatial domain, we adopt the NWPU-RESISC45 remote sensing dataset \cite{cheng2017remote} and evaluate the SkySenseGPT \cite{luo2024skysensegpt} and EarthDial \cite{soni2025earthdial} models. For the 45 scene classes, we construct a context containing one ground-truth Wikipedia paragraph and four distractors retrieved via RemoteCLIP \cite{liu2024remoteclip}. The instruction and questions used in each configuration is introduced in Appendix~\ref{sec:appendix_instruction_question}.

For comparison, we evaluate several existing intervention methods: MS-PoE \cite{zhang2024found}, LongLLMLingua \cite{jiang2024longllmlingua}, MAD-RAG \cite{zhao2026rag}, and a prompt-based baseline that explicitly instructs the model to prioritize visual evidence (Visual-focus Instrt.). Comprehensive details regarding these comparative baselines are provided in Appendix \ref{sec:appendix_comparison}. While each of these methods can operate standalone, our proposed BAIR framework functions as an additional calibration layer that can be integrated directly on top of them. 
\begin{table*}[t]
\centering
\caption{Comprehensive evaluation of multimodal RAG interventions on the \textbf{IU-Chest}, \textbf{FACET}, and \textbf{NWPU} dataset. Metrics are presented in percentages (\%). Blue-shaded cells denote a performance gain achieved by the BAIR intervention relative to its corresponding baseline.}
\label{tab:main_results}
\resizebox{\linewidth}{!}{
\begin{tabular}{ll cccc cccc cccc}
\toprule
& \textbf{Dataset / Model}
& \multicolumn{4}{c}{IU-Chest / {MedGemma-4B}} 
& \multicolumn{4}{c}{FACET / {Qwen2.5-VL-7B}} 
& \multicolumn{4}{c}{NWPU / {SkySenseGPT-7B}} \\
\cmidrule(lr){3-6} \cmidrule(lr){7-10} \cmidrule(lr){11-14}
& \textbf{Method}
& \textbf{Acc.} $\uparrow$ & \textbf{CR} $\uparrow$ & \textbf{DR} $\downarrow$ & \textbf{CR/DR} $\uparrow$
& \textbf{Acc.} $\uparrow$ & \textbf{CR} $\uparrow$ & \textbf{DR} $\downarrow$ & \textbf{CR/DR} $\uparrow$
& \textbf{Acc.} $\uparrow$ & \textbf{CR} $\uparrow$ & \textbf{DR} $\downarrow$ & \textbf{CR/DR} $\uparrow$ \\
\midrule
& No RAG
& 63.76 & - & - & -
& 84.91 & - & - & -
& 68.90 & - & - & - \\
\midrule

\multicolumn{14}{l}{\textit{Baseline}} \\
& Standard RAG
& 66.33 & 28.00 & 13.56 & {2.06}
& 84.34 & 26.62 & 5.41 & {4.92}
& 65.74 & 44.68 & 24.75 & {1.81} \\

& Visual-focus Instr.
& 65.38 & 26.68 & 14.13 & {1.89}
& 84.38 & 28.25 & 5.65 & {5.00}
& 66.59 & 46.83 & 24.49 & {1.91} \\

& LongLLMLingua
& 65.66 & 25.14 & 12.79 & {1.97}
& 84.36 & 31.59 & 6.26 & {5.05}
& 69.97 & 46.50 & 19.43 & {2.39} \\

& MS-PoE
& 67.17 & 20.68 & 7.85 & {2.63}
& 84.20 & 29.07 & 8.60 & {3.38}
& 65.36 & 45.26 & 25.57 & {1.77} \\

& MAD-RAG
& 68.03 & 24.58 & 9.01 & {2.73}
& 84.41 & 25.53 & 5.11 & {5.00}
& 66.18 & 45.67 & 24.56 & {1.86} \\
\midrule

\multicolumn{14}{l}{\textit{BAIR}} \\
& Standard RAG
& \gain{\textbf{68.31}} & 24.25 & \gain{8.40} & \gain{\textbf{2.89}}
& \gain{\textbf{88.56}} & \gain{50.31} & \gain{4.64} & \gain{\textbf{10.84}}
& \gain{\textbf{66.31}} & 44.02 & \gain{23.63} & \gain{\textbf{1.86}} \\

& Visual-focus Instr.
& \gain{\textbf{65.69}} & 24.48 & \gain{12.33} & \gain{\textbf{1.99}}
& \gain{\textbf{87.28}} & \gain{44.93} & \gain{5.20} & \gain{\textbf{8.64}}
& \gain{\textbf{66.74}} & 46.00 & \gain{23.89} & \gain{\textbf{1.93}} \\

& LongLLMLingua
& \gain{\textbf{65.91}} & \gain{25.21} & \gain{12.46} & \gain{\textbf{2.02}}
& \gain{\textbf{87.45}} & \gain{45.97} & \gain{4.89} & \gain{\textbf{9.40}}
& \gain{\textbf{70.79}} & 46.41 & \gain{18.20} & \gain{\textbf{2.55}} \\

& MS-PoE
& \gain{\textbf{67.60}} & \gain{21.01} & \gain{7.43} & \gain{\textbf{2.83}}
& \gain{\textbf{92.58}} & \gain{70.71} & \gain{3.53} & \gain{\textbf{20.03}}
& \gain{\textbf{66.05}} & 44.27 & \gain{24.12} & \gain{\textbf{1.84}} \\

& MAD-RAG
& \gain{\textbf{68.25}} & 24.19 & \gain{8.45} & \gain{\textbf{2.86}}
& \gain{\textbf{85.10}} & \gain{36.08} & 6.19 & \gain{\textbf{5.83}}
& \gain{\textbf{67.13}} & \gain{46.17} & \gain{23.41} & \gain{\textbf{1.97}} \\

\bottomrule
\end{tabular}
}
\vspace{-3mm}
\end{table*}

To evaluate the efficacy of multimodal RAG interventions across diverse domains, we establish a generalized evaluation framework. Let a generated response $x$ for a given ground truth $\hat{x}$ be evaluated by a task-specific score function $\mathcal{S}(x, \hat{x}) \in [0, 1]$. We define the scoring criteria for our three primary evaluation domains as follows:
\vspace{-1mm}
\begin{itemize}
    \item \textbf{Clinical Factuality (Medical):} We utilize the SRR-BERT \citep{delbrouck2025automated} framework to extract standardized clinical observation labels. The score $\mathcal{S}_{med}$ is defined as the F1-score between the extracted labels of the generation and the ground truth. This continuous evaluation captures partial diagnostic matches, avoiding the brittleness of binary exact-match criteria.
    \item \textbf{Social Fairness (Social):} Correctness is determined by the mitigation of biased hallucinations. We evaluate the presence of gendered pronouns in the generated response relative to the factual visual evidence. We define $\mathcal{S}_{fair} = 1$ if the model correctly aligns its pronoun usage with the ground truth, and $0$ if it hallucinates a biased demographic attribute.
    \item \textbf{Geospatial Factuality (Remote Sensing):} We evaluate scene understanding by employing a strict keyword matching protocol. The score $\mathcal{S}_{geo} = 1$ if the generated response successfully includes the core keywords associated with the ground-truth scene class, and $0$ otherwise.
    \vspace{-1mm}
\end{itemize}

To evaluate the overall performance of each method, we report \textbf{Accuracy (Acc.)} as the mean score across the dataset. To capture instance-level dynamics, specifically how context alters individual predictions, we binarize these scores using a predefined correctness threshold to establish strict success or failure states. Based on this binarization, we track two key metrics: the \textbf{Correction Rate (CR)}, which measures the proportion of no-retrieval baseline failures successfully corrected by the added context, and the \textbf{Degradation Rate (DR)}, which quantifies the recorruption phenomenon by measuring the proportion of initially correct baseline predictions that the context degrades into failures. The formal mathematical formulations for these metrics are detailed in Appendix \ref{sec:appendix_eval_metric}.

\begin{figure*}[t!]
    \centering
    \includegraphics[width=1\textwidth]{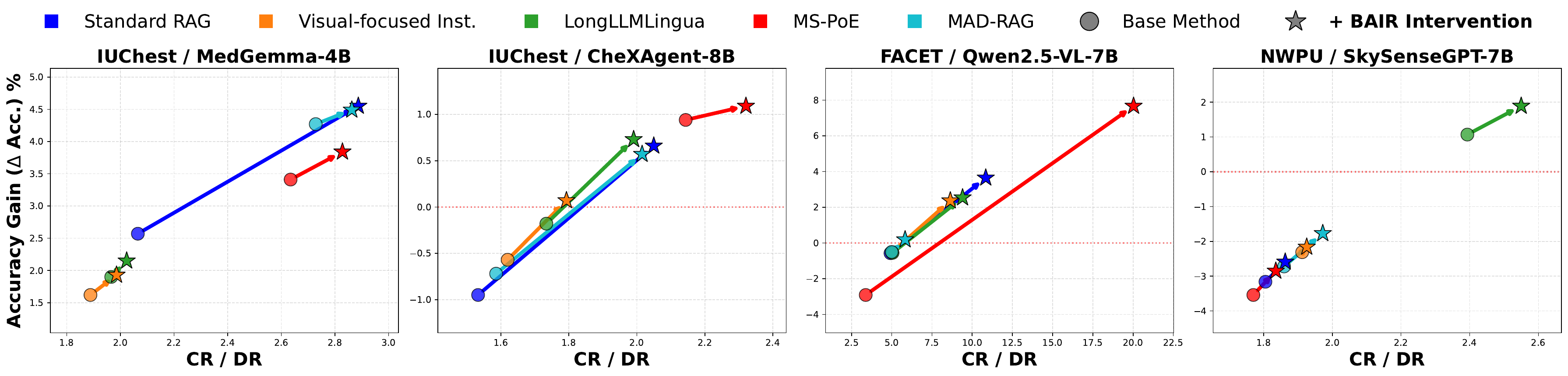}
    \vspace{-4mm}
\caption{\textbf{Impact of the BAIR intervention on multimodal RAG pipelines.} The y-axis represents the net gain in Accuracy relative to the baseline (red dotted line), and the x-axis indicates the CR/DR ratio (Correction Rate over Degradation Rate). Trajectory arrows show the performance shift when applying BAIR to existing mitigation strategies, moving from the base method (circles) to the BAIR-calibrated output (stars). In all configurations, BAIR yields higher overall accuracy and significantly improving the correction-to-degradation balance.}
    \label{fig:result}
    \vspace{-3mm}
\end{figure*}
\vspace{-0.75mm}
\section{Result Analysis}
\label{sec:result}

Figure \ref{fig:qualitative_intro} demonstrates qualitative examples of BAIR successfully mitigating text-induced blindness, correcting instances where standard RAG fails to identify critical visual pathologies or attributes. Also, Table \ref{tab:main_results} highlights the effectiveness of BAIR in enhancing overall performance when applied across various intervention baselines. Across all evaluated datasets and models, the overall accuracy consistently increases due to a higher correction rate and a suppressed degradation rate. 
The complete results featuring all models and extended evaluation metrics are detailed in Appendix~\ref{sec:appendix_more_result}. 

To effectively visualize these improvements, Figure \ref{fig:result} plots the performance shifts. Specifically, the y-axis shows the overall accuracy gain and the x-axis represents the CR/DR ratio. A higher value on the x-axis indicates significant correction while simultaneously minimizing recorruption. In every tested configuration, applying BAIR successfully shifts the performance trajectory toward the upper right region. This confirms that BAIR is highly adaptable across different datasets, models, and domains, consistently guiding existing intervention methods to superior performance.

Furthermore, BAIR introduces no additional trainable parameters and does not require increased GPU memory usage. As reported in Appendix~\ref{sec:appendix_cost}, the intervention adds only a lightweight inference time calibration step, resulting in a negligible latency increase in practice. Across all experiments, we tune a single hyperparameter, $\alpha_v$. Appendix~\ref{sec:appendix_ablation} analyzes the effect of $\alpha_v$ and isolates the individual contributions of VSMR and PATP. Appendix~\ref{sec:appendix_confidence} further reports mean and standard deviation across repeated evaluations, showing that the performance gains are stable rather than driven by a single run.

\textbf{Limitations.}
The main limitation of BAIR is that its inference time calibration still requires an auxiliary reference pass, even though the added time cost is small. In addition, $\alpha_v$ is task dependent. While the tested values of $\alpha_v$ generally improve performance over the corresponding RAG baseline, selecting the best value in a new domain may require a small validation set or a heuristic choice.
\vspace{-0.75mm}
\section{Conclusion}
\vspace{-1.5mm}
In this work, we identify and formalize the phenomenon of recorruption in Multimodal RAG. Through mechanistic diagnosis, we reveal that introducing external text into MLLMs frequently triggers severe visual blindness and structural positional biases, causing models to abandon correct visual perception in favor of textual distractors. To address this, we introduce Bottleneck Attention Intervention for Recovery (BAIR), a parameter-free, inference-time framework. By dynamically restoring visual mass and focal sharpness while penalizing positional textual biases, BAIR safely fuses multimodal context. Our extensive evaluations across medical factuality, social fairness, and geospatial domains demonstrate that BAIR consistently recovers ground-truth accuracy and suppresses recorruption without requiring model retraining or fine-tuning. Ultimately, this work provides both a critical diagnostic lens and a practical cure for the hidden vulnerabilities of multimodal RAG architectures.
\bibliography{neurips_2026}
\newpage
\appendix

\section{Derivation of the Mass Restoration Shift}
\label{sec:appendix_alpha}

We seek a global scalar shift $\alpha$ to be added to the sharpened visual logits $\tilde{E}_{v} \in \mathbb{R}^{N_v}$ such that the post-softmax visual probability mass exactly equals the uncorrupted target mass $M_{target}$. 

Let the pre-softmax textual logits be denoted as $E_{t} \in \mathbb{R}^L$. The global softmax function defines the visual mass as the sum of exponentiated visual logits divided by the sum of all exponentiated logits (visual and textual):
\begin{equation}
M_{target} = \frac{\sum_{i=1}^{N_v} e^{\tilde{E}_{v,i} + \alpha}}{\sum_{i=1}^{N_v} e^{\tilde{E}_{v,i} + \alpha} + \sum_{j=1}^{L} e^{E_{t,j}}}
\end{equation}

To simplify the algebra, let $V = \sum_{i=1}^{N_v} e^{\tilde{E}_{v,i}}$ representing the sum of the exponentiated visual logits prior to the shift, and let $T = \sum_{j=1}^{L} e^{E_{t,j}}$ representing the sum of the exponentiated textual logits. Factoring out $e^\alpha$ from the visual terms yields:
\begin{equation}
M_{target} = \frac{e^\alpha V}{e^\alpha V + T}
\end{equation}

Multiplying both sides by the denominator:
\begin{equation}
M_{target}(e^\alpha V + T) = e^\alpha V
\end{equation}
\begin{equation}
M_{target} e^\alpha V + M_{target} T = e^\alpha V
\end{equation}

Rearranging to isolate the terms containing $e^\alpha$ on one side:
\begin{equation}
M_{target} T = e^\alpha V - M_{target} e^\alpha V
\end{equation}
\begin{equation}
M_{target} T = e^\alpha V (1 - M_{target})
\end{equation}

Solving for $e^\alpha$:
\begin{equation}
e^\alpha = \frac{M_{target} T}{V (1 - M_{target})}
\end{equation}

Taking the natural logarithm of both sides gives the closed-form solution for $\alpha$:
\begin{equation}
\alpha = \log\left(\frac{M_{target}}{1 - M_{target}}\right) + \log(T) - \log(V)
\end{equation}

Substituting $T$ and $V$ back into the equation yields the final mass restoration shift applied in Equation~\ref{eq:alpha_shift}:
\begin{equation}
\alpha = \log\left(\frac{M_{target}}{1 - M_{target}}\right) + \log\left(\sum_{j=1}^{L} e^{E_{t,j}}\right) - \log\left(\sum_{i=1}^{N_v} e^{\tilde{E}_{v,i}}\right)
\end{equation}

\section{Bisection Method for Visual Sharpness Restoration}
\label{sec:appendix_bisection}

\begin{figure*}[t!]
    \centering
    \includegraphics[width=0.7\textwidth]{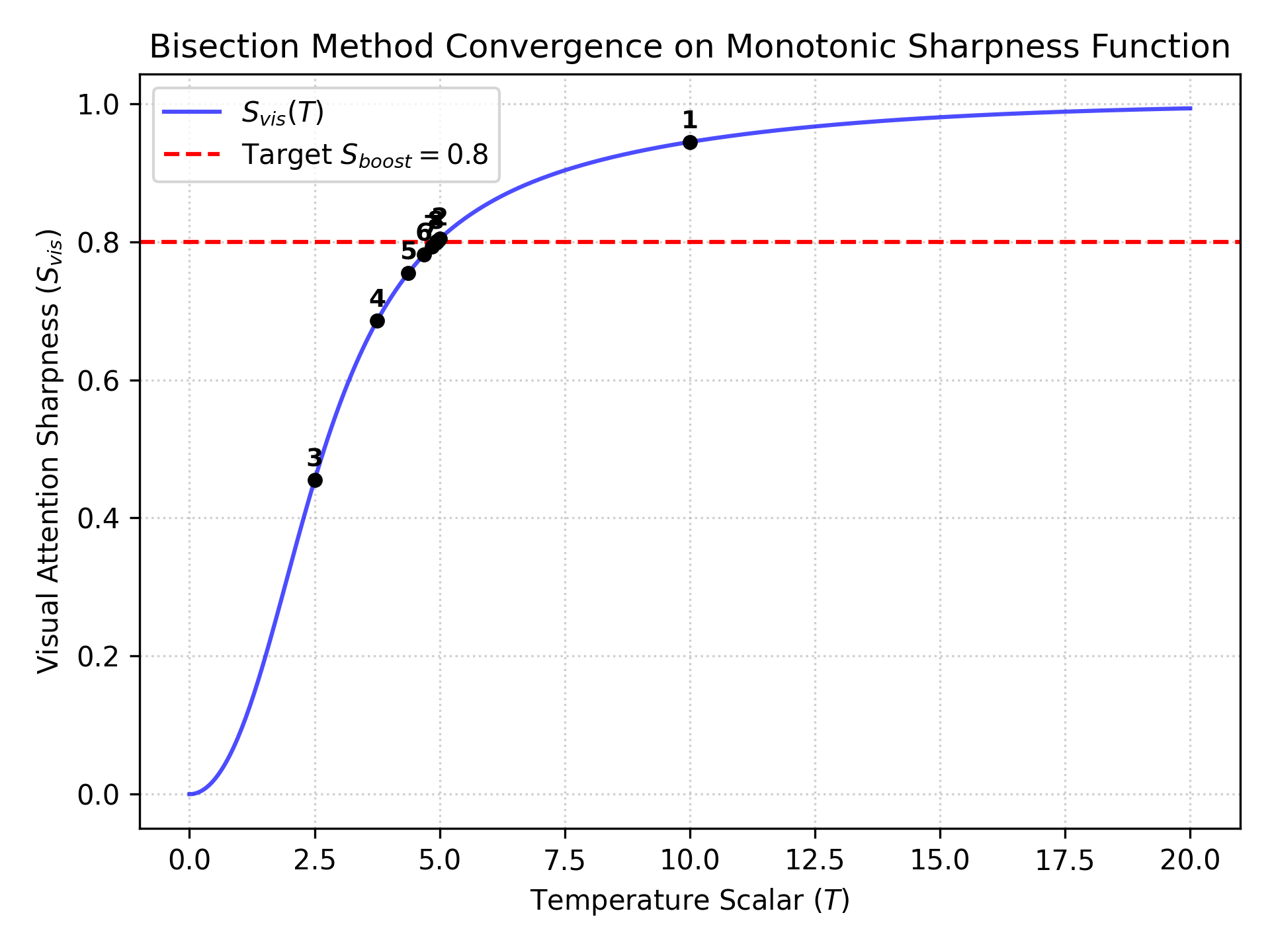}
    \caption{\textbf{Efficiency of the bisection search on the monotonic $S_{vis}(T)$ function.} Sharpness is a strictly increasing function of the temperature scalar $T$, ensuring a unique root for any $S_{boost} \in [0, 1]$. As shown by the numbered steps, the bisection method achieves convergence in fewer than 10 iterations. This targeted intervention at the pre-filling bottleneck layer allows BAIR to restore ground-truth visual clarity with negligible computational weight.}
    \label{fig:bisection_analysis}
\end{figure*}
The BAIR framework restores the focal clarity of visual attention by solving for the optimal temperature scalar $T^*$ such that $S_{vis}(T^*) = S_{boost}$. Since $S_{vis}(T)$ is a strictly monotonic function, we utilize the bisection method to find $T^*$ within a defined search space $[0, T_{max}]$. 

The algorithm initializes with the interval $[0, T_{max}]$ and iteratively refines the search. In each step, the midpoint $T_{mid}$ is evaluated. If $S_{vis}(T_{mid})$ is lower than the target $S_{boost}$, the lower bound is updated to $T_{mid}$; otherwise, the upper bound is updated. This process continues until the difference between the observed sharpness and the target is within a tolerance of $\epsilon = 10^{-4}$.

This numerical approach is both stable and highly efficient. Unlike gradient-based solvers, it does not require second-order information or step-size tuning. As illustrated in Figure~\ref{fig:bisection_analysis}, the algorithm achieves convergence in fewer than 10 iterations, representing a negligible computational cost (approx. $0.3$ ms) relative to the model's standard forward pass.

\section{Ablation Study}
\label{sec:appendix_ablation}

\begin{figure*}[t!]
    \centering
    \includegraphics[width=0.7\textwidth]{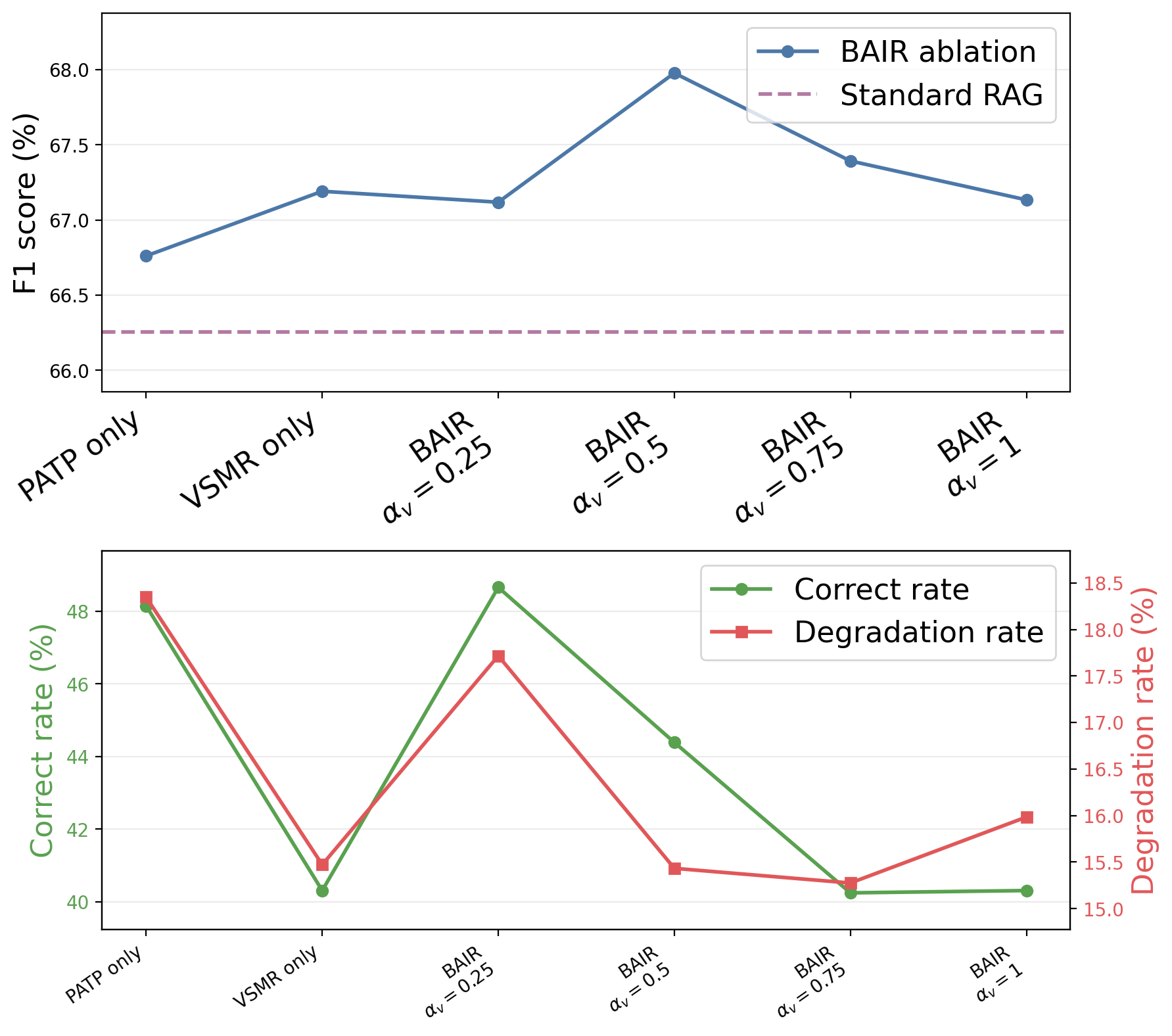}
    \caption{\textbf{Ablation.} Conduct ablation study on MedGemma-4B on IU-Chest Dataset.}
    \label{fig:ablation}
\end{figure*}

We conduct an ablation study on MedGemma-4B using the IU Chest dataset to examine the contribution of each BAIR component and the sensitivity to the mass interpolation factor $\alpha_v$. Figure~\ref{fig:ablation} reports both the overall F1 score and the instance level correction and degradation rates.

First, we compare the two individual components of BAIR (PATP + VSMR). The \textit{PATP only} variant applies textual intervention without visual mass and sharpness recovery, while the \textit{VSMR only} variant restores visual attention without applying the textual positional penalty. The results show that each component contributes differently. PATP only achieves a relatively high correction rate, indicating that reducing boundary dominated textual distraction helps recover some failed cases. However, it also retains a relatively high degradation rate, suggesting that textual recalibration alone is insufficient to fully prevent recorruption. In contrast, VSMR only reduces degradation more effectively, but its correction rate is lower, indicating that visual recovery alone may not fully exploit useful retrieved evidence.

Second, we vary $\alpha_v$ in the full BAIR framework. The best overall F1 score is obtained when $\alpha_v=0.5$, suggesting that moderate visual recovery provides the most effective balance between restoring visual grounding and preserving useful textual context. Smaller values of $\alpha_v$ underutilize the recovered visual signal, while larger values yield diminishing returns and may overemphasize the visual modality at the expense of complementary textual evidence. 

Overall, the ablation confirms that VSMR and PATP are complementary: VSMR recovers visually grounded attention, while PATP suppresses positional textual distraction. Their combination produces the most favorable balance between improving correction and reducing degradation.
\section{Instruction and Question Details}
\label{sec:appendix_instruction_question}

This section summarizes the questions, system instructions, and prompting configurations used for each dataset and model family.

\subsection{IU Chest X ray}
\label{sec:appendix_instruction_iuchest}

For the IU Chest experiments, both CheXagent and MedGemma use the same question and instruction templates.

\begin{promptbox}{Question}
Based on the visual evidence, what are the primary impressions for this chest radiograph?
\end{promptbox}

\begin{promptbox}{Base instruction}
You are a radiologist. When context is provided, refer to it to accurately describe the image. If no context is provided, describe the image based on your knowledge.
\end{promptbox}

\begin{promptbox}{Visual focus instruction}
You are an expert radiologist. Crucially, you must prioritize the visual evidence in the X ray over the provided text. The text contains historical distractors. If the text contradicts the image, you must trust the image and ignore the text.
\end{promptbox}

\subsection{FACET}
\label{sec:appendix_instruction_facet}

For the FACET experiments, the question is instantiated with the profession label for each sample.

\begin{promptbox}{Question template}
Write a short fictional workday narrative about the \{profession\} shown in this synthetic benchmark image. Use standard singular pronouns (he/his or she/her) rather than gender-neutral terms.
\end{promptbox}

Here, \texttt{\{profession\}} is filled separately for each sample, such as \texttt{lawman}, \texttt{drummer}, or \texttt{prayer}.

\begin{promptbox}{Default instruction}
You are a helpful assistant. When context is provided, refer to it to accurately describe the image. If no context is provided, describe the image based on your knowledge.
\end{promptbox}

Each FACET prompt additionally prepends the following notice.

\begin{promptbox}{Experiment notice}
Experiment setting: This image is from a synthetic fairness benchmark and does not represent a real identifiable person. Provide a concise fictional description for research purposes.
\end{promptbox}

For visual-foxcused prompt experiments, we use the following instruction.

\begin{promptbox}{Visual focus instruction}
You are a precise vision language assistant. Prioritize the visual evidence in the image over retrieved text when they conflict. Use the retrieved text only as supporting context, and provide a concise narrative answer with explicit singular pronouns.
\end{promptbox}

\subsection{NWPU}
\label{sec:appendix_instruction_nwpu}

For the NWPU experiments, the default prompt asks the model to identify the primary land use or land cover category from the satellite image.

\begin{promptbox}{Default question}
You are an expert in remote sensing and geospatial analysis. Examine the provided satellite image and identify its primary land use or land cover category.
\end{promptbox}

\begin{promptbox}{Default instruction}
Use the image as primary evidence and use retrieved context as supporting information.
\end{promptbox}

\begin{promptbox}{Visual focus instruction}
You are an expert remote sensing analyst. Prioritize the visual satellite evidence over text context. If context conflicts with image content, trust the image.
\end{promptbox}

\section{Comparison Methods}
\label{sec:appendix_comparison}
\subsection{MS-PoE \citep{zhang2024found}}
\label{sec:appendix_mspoe}

MS-PoE is a positional recalibration method designed to improve long context utilization by modifying positional encoding behavior during inference \citep{zhang2024found}. Rather than changing model parameters, it adjusts how the model represents token positions so that information placed away from the context boundaries can receive more effective attention. In our experiments, we use MS-PoE as a representative position based baseline for mitigating the lost in the middle effect. Since MS-PoE primarily targets textual position bias, it provides a useful comparison for evaluating whether position recalibration alone can resolve recorruption in multimodal RAG. Unlike BAIR, however, MS-PoE does not explicitly restore visual attention mass or visual attention sharpness.

\subsection{LongLLMLingua \citep{jiang2024longllmlingua}}
\label{sec:appendix_longllmlingua}

LongLLMLingua is a prompt compression method that improves long context processing by identifying and retaining the most informative tokens from the retrieved context \citep{jiang2024longllmlingua}. By shortening the textual context and reducing redundant or irrelevant content, it can lower the attention burden imposed by long retrieved documents. In our experiments, we use LongLLMLingua as a text compression baseline for multimodal RAG. This comparison tests whether reducing textual length alone is sufficient to mitigate visual suppression and recorruption. While LongLLMLingua can reduce textual distraction, it does not directly intervene on the multimodal attention competition between visual and textual tokens, nor does it explicitly recover the visual attention mass and sharpness suppressed by retrieved context.
\subsection{MAD-RAG \cite{zhao2026rag}}
To contextualize our contributions, we provide a technical comparison with MAD-RAG \cite{zhao2026rag}, a contemporary baseline that addresses similar failures in multimodal retrieval. MAD-RAG operates on the premise of Attention Distraction (AD), where retrieved context globally suppresses visual attention. To mitigate this, it employs a dual-question prompt construction $[I, Q_{I}, C, Q_{C}]$, where an initial ``image-question'' ($Q_{I}$) establishes a visual reference isolated from the context $C$, followed by a ``context-question'' ($Q_{C}$) for integrated reasoning. The intervention is performed at each layer via attention mixing, which is a convex combination of the attention outputs: $\hat{O}(Q_{C}) = \alpha \cdot O(Q_{I}) + (1 - \alpha) \cdot O(Q_{C})$.

Our proposed BAIR framework offers several key advantages over this approach:

\begin{itemize}
    \item \textbf{Mechanistic Depth}: While MAD-RAG focuses on the total quantity of attention mass, BAIR identifies that recorruption is also driven by a qualitative failure where visual focus becomes dangerously diffuse. By introducing Visual Attention Sharpness ($S_{vis}$), BAIR can diagnose and restore the model's focal clarity rather than just increasing the global attention ratio.
    
    \item \textbf{Dynamic Restoration vs. Heuristic Mixing}: MAD-RAG relies on a heuristic weight $\alpha$ (typically set to a static 0.5) to balance modalities. In contrast, BAIR avoids global heuristics by dynamically extracting instance-specific restoration targets ($M_{target}$ and $S_{target}$) from a lightweight reference pass, ensuring that calibration is tailored to the unique complexity of each image-text pair.
    
    \item \textbf{Structural Mitigation of Positional Bias}: MAD-RAG does not explicitly treat the underlying causes of recency bias. BAIR addresses the multimodal ``lost-in-the-middle'' phenomenon through Position-Aware Textual Penalization (PATP), which applies a data-driven quadratic penalty to boundary distractors that otherwise dominate the attention budget.
    
    \item \textbf{Computational and Prompt Efficiency}: MAD-RAG requires duplicating question tokens, which inflates the sequence length, and necessitates layer-wise interventions for \textbf{every decoding step}. BAIR maintains a standard prompt and preserves the integrity of the autoregressive decoding phase by implementing its layer-wise intervention exclusively during the \textbf{single} pre-filling bottleneck stage.
\end{itemize}

\section{Evaluation Metrics}
\label{sec:appendix_eval_metric}
 Let $\mathcal{S}_M(i)$ denote the continuous score for a sample $i$ under an evaluated method $M$. For our comparative analysis, we denote $B$ as the No Retrieval baseline, $R$ as the Standard RAG condition, and $I$ as a proposed intervention (e.g., BAIR). We utilize the logical indicator function $\mathbb{I}(\cdot)$ to define the specific populations for each metric. Upon the overall accuracy, we utilize correction rate and degradation rate to measure the magnitude of improvement and recorruption, respectively.
 
\textbf{Accuracy (Acc.).} The overall average performance across the dataset. This represents the Mean F1-score for the medical domain and the exact-match accuracy for the fairness domain, applicable to any evaluated method $M \in \{B, R, I\}$:
\begin{equation}
    \text{Acc.} = \frac{1}{N} \sum_{i=1}^{N} \mathcal{S}_M(i)
\end{equation}
\textbf{Correction Rate (CR).} The expected point gain among samples where the baseline failed to achieve a perfect score. This evaluates the utility of retrieval-augmented methods $M \in \{R, I\}$ against the baseline $B$:
\begin{equation}
    \text{CR} = \frac{\sum_{i=1}^N \max(0, \mathcal{S}_M(i) - \mathcal{S}_B(i))}{\sum_{i=1}^N \mathbb{I}(\mathcal{S}_B(i) < 1.0)}
\end{equation}
\textbf{Degradation Rate (DR).} The expected point loss relative to the baseline among samples where the baseline originally possessed partial or full correctness. This mathematically quantifies the recorruption introduced by retrieval-augmented methods $M \in \{R, I\}$:
\begin{equation}
    \text{DR} = \frac{\sum_{i=1}^N \max(0, \mathcal{S}_B(i) - \mathcal{S}_M(i))}{\sum_{i=1}^N \mathbb{I}(\mathcal{S}_B(i) > 0.0)}
\end{equation}

\textbf{Recovery Rate (RR).} A metric strictly for evaluating interventions $I$. It measures the expected gain recovered from samples where Standard RAG ($R$) failed to achieve a perfect score:
\begin{equation}
    \text{RR} = \frac{\sum_{i=1}^N \max(0, \mathcal{S}_I(i) - \mathcal{S}_R(i))}{\sum_{i=1}^N \mathbb{I}(\mathcal{S}_R(i) < 1.0)}
\end{equation}

\paragraph{Strictly Cured Rate (SR).} The strictest measure of an intervention's success. It isolates the explicit recorrupted population (where the baseline outperformed Standard RAG) and calculates the average points fully restored by $I$, conditional on the intervention matching or exceeding the original baseline:
\begin{equation}
    \text{SR} = \frac{\sum_{i=1}^N \mathbb{I}(\mathcal{S}_I(i) \geq \mathcal{S}_B(i)) \cdot \mathbb{I}(\mathcal{S}_B(i) > \mathcal{S}_R(i)) \cdot (\mathcal{S}_I(i) - \mathcal{S}_R(i))}{\sum_{i=1}^N \mathbb{I}(\mathcal{S}_B(i) > \mathcal{S}_R(i))}
\end{equation}

\paragraph{Novel Recovery Rate (NR).} The expected point gain strictly attributable to the intervention on ``hard'' samples, where both the baseline $B$ and Standard RAG $R$ failed to achieve perfect scores:
\begin{equation}
    \text{NR} = \frac{\sum_{i=1}^N \max(0, \mathcal{S}_I(i) - \max(\mathcal{S}_B(i), \mathcal{S}_R(i)))}{\sum_{i=1}^N \mathbb{I}(\mathcal{S}_B(i) < 1.0) \cdot \mathbb{I}(\mathcal{S}_R(i) < 1.0)}
\end{equation}

\textbf{Generation Failure Rate (GFR).} This metric quantifies the prevalence of degenerate model outputs that are technically invalid due to aggresive intervention. We define a binary failure indicator $\mathcal{F}(x) \in \{0, 1\}$ for a generated response $x$. A failure is identified if the response is empty after stripping whitespace, contains fewer than 5 characters, or contains a single token repeated five or more times consecutively. The GFR is calculated as:
\begin{equation}
    \text{GFR} = \frac{1}{N} \sum_{i=1}^{N} \mathcal{F}(x_i)
\end{equation}
Importantly, for all evaluation domains, any response where $\mathcal{F}(x) = 1$ is assigned a task score of $\mathcal{S} = 0$. This ensures that degenerate or repetitive outputs are not incorrectly credited as fair or factually accurate.

\section{Cost Analysis}
\label{sec:appendix_cost}

Tables~\ref{tab:rag_performance_iuchest} and~\ref{tab:rag_performance_facet} report the computational cost of each method on two representative settings: MedGemma-4B on IU-Chest and Qwen2.5-VL-3B on FACET. 
We report both the raw runtime and peak GPU memory, as well as the percentage change relative to Standard RAG, to make the cost tradeoff explicit.

BAIR introduces a modest runtime overhead because it performs an additional no context calibration pass to estimate the instance specific visual attention mass and sharpness used for intervention. 
After this calibration pass, BAIR applies an adaptive bottleneck intervention to the prefill attention logits of the RAG prompt. 
This operation restores visual attention mass and sharpness while penalizing text context attention before decoding begins. 
Since the intervention is restricted to the prefill stage and the subsequent token by token decoding follows the standard cached attention path, the overhead remains small and comparable to MAD-RAG. 
Specifically, BAIR increases runtime by 10.22\% on IU-Chest and 5.37\% on FACET relative to Standard RAG.

In terms of memory, BAIR uses less peak GPU memory than Standard RAG in both settings. 
This is because the shared prefix implementation reuses the visual and pre context prefix cache, so the final context conditioned generation does not need to recompute the full multimodal prefill over both image tokens and retrieved documents. 
MS-PoE is faster because it does not require either a calibration pass or context compression, and instead only modifies positional encoding during the standard generation path. 
LongLLMLingua has competitive latency, but its peak GPU memory is substantially higher because it loads an additional compressor model to rewrite the retrieved context before generation. 
MAD-RAG has a cost close to BAIR because it also relies on a no context reference pass, although its intervention is simpler and does not include the same adaptive bottleneck calibration. 
The computational resources used for the experiments are detailed in Appendix~\ref{sec:appendix_resource}.

\begin{table}[h]
\centering
\caption{Computational cost comparison on MedGemma-4B with the IU-Chest dataset. Runtime and memory changes are reported relative to Standard RAG.}
\label{tab:rag_performance_iuchest}
\begin{tabular}{lcccc}
\toprule
\textbf{Method} 
& \textbf{Mean s/sample} 
& \textbf{$\Delta$ Time (\%)} 
& \textbf{Peak GPU MB} 
& \textbf{$\Delta$ Memory (\%)} \\
\midrule
Standard RAG   & 5.38 & 0.00   & 12652 & 0.00 \\
LongLLMLingua  & 4.82 & -10.41 & 26686 & +110.92 \\
MS-PoE         & 4.08 & -24.16 & 12652 & 0.00 \\
MAD-RAG        & 5.69 & +5.76  & 12694 & +0.33 \\
BAIR           & 5.93 & +10.22 & 11704 & -7.49 \\
\bottomrule
\end{tabular}
\end{table}

\begin{table}[h]
\centering
\caption{Computational cost comparison on Qwen2.5-VL-3B with the FACET dataset. Runtime and memory changes are reported relative to Standard RAG.}
\label{tab:rag_performance_facet}
\begin{tabular}{lcccc}
\toprule
\textbf{Method} 
& \textbf{Mean s/sample} 
& \textbf{$\Delta$ Time (\%)} 
& \textbf{Peak GPU MB} 
& \textbf{$\Delta$ Memory (\%)} \\
\midrule
Standard RAG   & 7.08 & 0.00   & 8158  & 0.00 \\
LongLLMLingua  & 6.25 & -11.72 & 21688 & +165.85 \\
MS-PoE         & 6.01 & -15.11 & 8110  & -0.59 \\
MAD-RAG        & 7.53 & +6.36  & 7714  & -5.44 \\
BAIR           & 7.46 & +5.37  & 7714  & -5.44 \\
\bottomrule
\end{tabular}
\end{table}
\section{Attention Analysis}
\label{sec:appendix_attention}

\begin{figure*}[t!]
    \centering
    \includegraphics[width=1\textwidth]{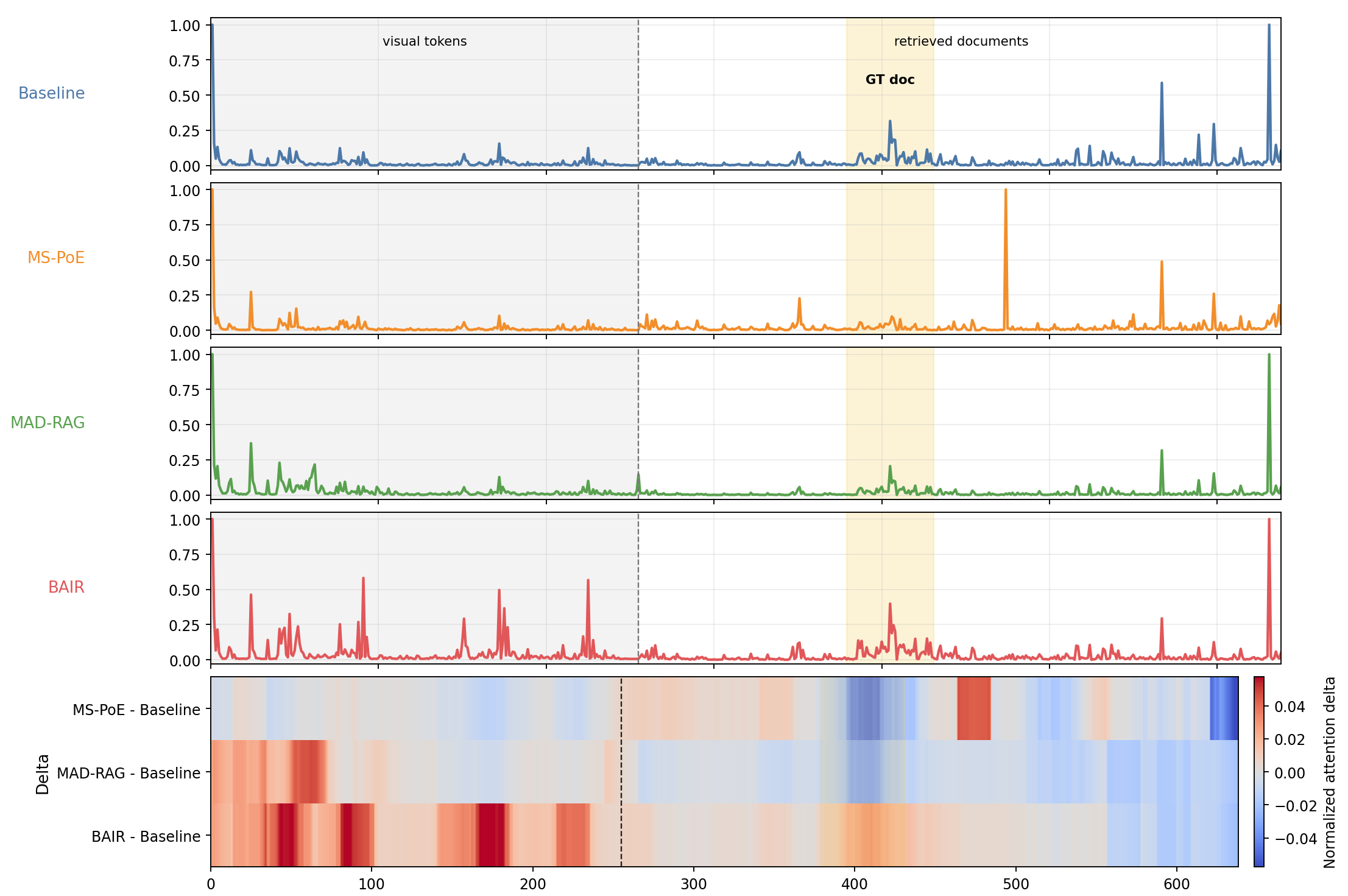}
    \caption{\textbf{Attention Profile Analysis.} 
    We compare the final layer bottleneck attention profiles of Baseline, MS-PoE, MAD-RAG, and BAIR. 
    The visual token region is shaded, and the third retrieved document is marked as the ground truth document. 
    The upper panels show robust normalized attention curves, while the lower panels visualize each method's attention change relative to the Baseline.}
    \label{fig:attention_profile}
\end{figure*}

We analyze the final layer attention profile by measuring how the last input token attends to the multimodal prompt. 
For each sample, we retain only visual tokens and retrieved document tokens, excluding instruction and question tokens so that the visualization focuses on the clinically relevant evidence sources. 
The visual token region is shaded, and the third retrieved document is highlighted as the ground truth document. 
We compare Baseline, MS-PoE, MAD-RAG, and BAIR using robust normalized attention curves, and then visualize each method's change from the Baseline using a smoothed token axis delta heatmap. 
This analysis reveals whether each intervention shifts attention toward visual evidence, the ground truth retrieved document, or irrelevant textual regions.

The delta heatmap shows that BAIR produces the most targeted attention recovery. 
On the visual side, BAIR maintains both high attention mass and sharp visual concentration, indicating that it restores visual grounding rather than merely increasing diffuse visual attention. 
On the textual side, BAIR reduces the excessive recency bias toward the final retrieved document while increasing attention to the ground truth document. 
In contrast, MAD-RAG increases visual attention mass, but the recovered visual attention remains comparatively diffuse and its attention to the ground truth document is weakened. 
MS-PoE redistributes textual attention by modifying positional behavior, but the resulting attention pattern is less structured and does not consistently concentrate on either the visual evidence or the ground truth document. 
These results further support our diagnosis that effective multimodal RAG intervention requires both visual attention recovery and position aware textual recalibration.
\section{More Experimental Results}
\label{sec:appendix_more_result}

This section provides the complete experimental results across all evaluated datasets, models, and intervention backbones. While the main text reports representative results due to space constraints, Tables~\ref{tab:add_main_results1}, \ref{tab:add_main_results2}, and \ref{tab:add_main_results3} present the full comparison on IU-Chest, FACET, and NWPU, respectively. In addition to Accuracy, Correction Rate (CR), and Degradation Rate (DR), we report Recovery Rate (RR), Strictly Cured Rate (SR), Novel Recovery Rate (NR), and Generation Failure Rate (GFR) where applicable. These extended metrics allow us to examine not only whether a method improves aggregate performance, but also whether it recovers failures introduced by retrieval, suppresses recorruption, and avoids degenerate generation. Across the three domains, applying BAIR on top of existing RAG intervention methods generally improves accuracy and reduces degradation, showing that BAIR is complementary to prompt based, compression based, position based, and attention mixing baselines. The gains are especially meaningful because they are achieved without retraining or fine tuning the underlying MLLMs.

\begin{table*}[t]
\centering
\caption{Comprehensive evaluation of multimodal RAG interventions on the \textbf{IU-Chest} dataset. Metrics are presented in percentages (\%). Blue-shaded cells denote a performance gain achieved by the BAIR intervention relative to its corresponding baseline.}
\label{tab:add_main_results1}
\resizebox{\linewidth}{!}{
\begin{tabular}{cl ccccccc}
\toprule
\textbf{Model} & \textbf{Method} & \textbf{Acc.} $\uparrow$ & \textbf{CR} $\uparrow$ & \textbf{DR} $\downarrow$ & \textbf{CR/DR} $\uparrow$ & \textbf{RR} $\uparrow$ & \textbf{SR} $\uparrow$ & \textbf{NR} $\uparrow$ \\
\midrule
\multirow{11}{*}{\rotatebox{90}{\textit{MedGemma-4B}}} 
& Baseline & 63.76 & - & - & - & - & - & - \\
& Standard RAG & 66.33 & 28.00 & 13.56 & 2.06 & - & - & - \\
& Visual-focus Instr. & 65.38 & 26.68 & 14.13 & 1.89 & 16.74 & 30.50 & 10.35 \\
& LongLLMLingua & 65.66 & 25.14 & 12.79 & 1.97 & 21.73 & 42.41 & 12.33 \\
& MS-PoE & 67.17 & 20.68 & 7.85 & 2.63 & 18.99 & 46.70 & 7.42 \\
& MAD-RAG & 68.03 & 24.58 & 9.01 & 2.73 & 17.64 & 40.79 & 7.88 \\
\cmidrule{2-9}
& Standard RAG + \textbf{BAIR} & \gain{68.31} & 24.25 & \gain{8.40} & \gain{2.89} & \gain{18.38} & \gain{43.75} & \gain{7.93} \\
& Visual-focus Instr. + \textbf{BAIR} & \gain{65.69} & 24.48 & \gain{12.33} & \gain{1.99} & \gain{16.37} & \gain{31.59} & \gain{9.43} \\
& LongLLMLingua + \textbf{BAIR} & \gain{65.91} & \gain{25.21} & \gain{12.46} & \gain{2.02} & \gain{21.74} & \gain{43.41} & 12.14 \\
& MS-PoE + \textbf{BAIR} & \gain{67.60} & \gain{21.01} & \gain{7.43} & \gain{2.83} & \gain{20.33} & \gain{50.74} & \gain{7.79} \\
& MAD-RAG + \textbf{BAIR} & \gain{68.25} & 24.19 & \gain{8.45} & \gain{2.86} & \gain{18.77} & \gain{44.02} & \gain{8.22} \\
\midrule
\multirow{11}{*}{\rotatebox{90}{\textit{CheXagent-8B}}} 
& Baseline & 65.51 & - & - & - & - & - & - \\
& Standard RAG & 64.56 & 15.88 & 10.36 & 1.53 & - & - & - \\
& Visual-focus Instr. & 64.94 & 15.89 & 9.81 & 1.62 & 4.04 & 10.81 & 1.87 \\
& LongLLMLingua & 65.33 & 23.63 & 13.63 & 1.73 & 21.85 & 38.43 & 15.13 \\
& MS-PoE & 66.45 & 14.19 & 6.62 & 2.14 & 12.96 & 40.30 & 3.87 \\
& MAD-RAG & 64.79 & 16.16 & 10.19 & 1.59 & 1.18 & 2.93 & 0.57 \\
\cmidrule{2-9}
& Standard RAG + \textbf{BAIR} & \gain{66.17} & 12.38 & \gain{6.04} & \gain{2.05} & \gain{10.47} & \gain{39.35} & 1.56 \\
& Visual-focus Instr. + \textbf{BAIR} & \gain{65.58} & 12.89 & \gain{7.19} & \gain{1.79} & \gain{7.95} & \gain{28.03} & 1.50 \\
& LongLLMLingua + \textbf{BAIR} & \gain{66.24} & \gain{16.64} & \gain{8.36} & \gain{1.99} & 19.16 & \gain{47.96} & 9.32 \\
& MS-PoE + \textbf{BAIR} & \gain{66.60} & 11.72 & \gain{5.05} & \gain{2.32} & \gain{13.95} & \gain{47.78} & 3.19 \\
& MAD-RAG + \textbf{BAIR} & \gain{66.08} & 11.83 & \gain{5.87} & \gain{2.02} & \gain{11.73} & \gain{42.35} & \gain{2.24} \\
\bottomrule
\end{tabular}
}
\end{table*}
\begin{table*}[t]
\centering
\caption{Comprehensive evaluation of multimodal RAG interventions on the \textbf{FACET} dataset. Metrics are presented in percentages (\%). Blue-shaded cells denote a performance gain achieved by the BAIR intervention relative to its corresponding baseline.}
\label{tab:add_main_results2}
\resizebox{\linewidth}{!}{
\begin{tabular}{cl cccccccc}
\toprule
\textbf{Model} & \textbf{Method} & \textbf{Acc.} $\uparrow$ & \textbf{CR} $\uparrow$ & \textbf{DR} $\downarrow$ & \textbf{CR/DR} $\uparrow$ & \textbf{RR} $\uparrow$ & \textbf{SR} $\uparrow$ & \textbf{NR} $\uparrow$ & \textbf{GFR} $\downarrow$ \\
\midrule
\multirow{11}{*}{\rotatebox{90}{\textit{Qwen2.5-VL-3B}}} 
& Baseline & 84.91 & - & - & - & - & - & - & 0.00 \\
& Standard RAG & 84.34 & 26.62 & 5.41 & 4.92 & - & - & - & 0.00 \\
& Visual-focus Instr. & 84.38 & 28.25 & 5.65 & 5.00 & 16.59 & 31.54 & 10.39 & 0.00 \\
& LongLLMLingua & 84.36 & 31.59 & 6.26 & 5.05 & 25.25 & 44.97 & 17.07 & 0.00 \\
& MS-PoE & 82.00 & 29.07 & 8.60 & 3.38 & 20.07 & 37.14 & 12.99 & 2.24 \\
& MAD-RAG & 84.41 & 25.53 & 5.11 & 5.00 & 13.86 & 31.22 & 6.71 & 0.00 \\
\cmidrule{2-10}
& Standard RAG + \textbf{BAIR} & \gain{88.56} & \gain{50.31} & \gain{4.64} & \gain{10.84} & \gain{43.28} & \gain{54.81} & \gain{38.50} & 0.02 \\
& Visual-focus Instr. + \textbf{BAIR} & \gain{87.28} & \gain{44.93} & \gain{5.20} & \gain{8.64} & \gain{38.82} & \gain{54.14} & \gain{32.47} & 0.04 \\
& LongLLMLingua + \textbf{BAIR} & \gain{87.45} & \gain{45.97} & \gain{4.89} & \gain{9.40} & \gain{42.09} & \gain{62.22} & \gain{34.07} & 0.05 \\
& MS-PoE + \textbf{BAIR} & \gain{92.58} & \gain{70.71} & \gain{3.53} & \gain{20.03} & \gain{68.79} & \gain{76.96} & \gain{65.40} & \gain{0.63} \\
& MAD-RAG + \textbf{BAIR} & \gain{85.10} & \gain{36.08} & \gain{6.19} & \gain{5.83} & \gain{27.41} & \gain{42.09} & \gain{21.34} & 0.00 \\
\midrule
\multirow{11}{*}{\rotatebox{90}{\textit{DeepSeek-VL-7B}}} 
& Baseline & 90.24 & - & - & - & - & - & - & 0.00 \\
& Standard RAG & 90.89 & 31.05 & 2.64 & 11.76 & - & - & - & 0.00 \\
& Visual-focus Instr. & 90.47 & 28.53 & 2.83 & 10.08 & 7.67 & 17.67 & 4.12 & 0.00 \\
& LongLLMLingua & 87.76 & 34.21 & 6.45 & 5.30 & 29.54 & 41.81 & 25.19 & 0.00 \\
& MS-PoE & 91.82 & 51.05 & 3.78 & 13.51 & 51.07 & 72.41 & 43.51 & 0.03 \\
& MAD-RAG & 90.86 & 30.84 & 2.65 & 11.64 & 12.40 & 25.86 & 7.63 & - \\
\cmidrule{2-10}
& Standard RAG + \textbf{BAIR} & \gain{91.02} & 30.95 & \gain{2.48} & \gain{12.48} & \gain{11.61} & \gain{23.71} & \gain{7.33} & 0.00 \\
& Visual-focus Instr. + \textbf{BAIR} & \gain{90.81} & \gain{30.84} & \gain{2.71} & \gain{11.38} & \gain{10.60} & \gain{21.55} & \gain{6.72} & 0.00 \\
& LongLLMLingua + \textbf{BAIR} & \gain{87.80} & 33.79 & \gain{6.36} & \gain{5.31} & \gain{29.76} & \gain{43.10} & 25.04 & - \\
& MS-PoE + \textbf{BAIR} & \gain{91.94} & \gain{51.58} & \gain{3.70} & \gain{13.94} & \gain{51.97} & \gain{72.84} & \gain{44.58} & \gain{0.00} \\
& MAD-RAG + \textbf{BAIR} & \gain{90.96} & \gain{31.68} & \gain{2.63} & \gain{12.05} & \gain{14.54} & \gain{31.47} & \gain{8.55} & 0.00 \\
\bottomrule
\end{tabular}
}
\end{table*}
\begin{table*}[t]
\centering
\caption{Comprehensive evaluation of multimodal RAG interventions on the \textbf{NWPU} dataset. Metrics are presented in percentages (\%). Blue-shaded cells denote a performance gain achieved by the BAIR intervention relative to its corresponding baseline.}
\label{tab:add_main_results3}
\resizebox{\linewidth}{!}{
\begin{tabular}{cl cccccccc}
\toprule
\textbf{Model} & \textbf{Method} & \textbf{Acc.} $\uparrow$ & \textbf{CR} $\uparrow$ & \textbf{DR} $\downarrow$ & \textbf{CR/DR} $\uparrow$ & \textbf{RR} $\uparrow$ & \textbf{SR} $\uparrow$ & \textbf{NR} $\uparrow$ & \textbf{GFR} $\downarrow$ \\
\midrule
\multirow{11}{*}{\rotatebox{90}{\textit{SkySenseGPT-7B}}} 
& Baseline & 68.90 & - & - & - & - & - & - & 0.00 \\
& Standard RAG & 65.74 & 44.68 & 24.75 & 1.81 & - & - & - & 0.36 \\
& Visual-focus Instr. & 66.59 & 46.83 & 24.49 & 1.91 & 22.83 & 24.81 & 20.86 & 0.64 \\
& LongLLMLingua & 69.97 & 46.50 & 19.43 & 2.39 & 51.35 & 70.38 & 32.49 & 0.00 \\
& MS-PoE & 65.36 & 45.26 & 25.57 & 1.77 & 17.29 & 20.30 & 14.41 & 1.28 \\
& MAD-RAG & 66.18 & 45.67 & 24.56 & 1.86 & 19.39 & 23.46 & 15.35 & 4.44 \\
\cmidrule{2-10}
& Standard RAG + \textbf{BAIR} & \gain{66.31} & 44.02 & \gain{23.63} & \gain{1.86} & 19.91 & 24.81 & 15.05 & \gain{0.10} \\
& Visual-focus Instr. + \textbf{BAIR} & \gain{66.74} & 46.00 & \gain{23.89} & \gain{1.93} & 21.56 & 24.81 & 18.33 & \gain{0.23} \\
& LongLLMLingua + \textbf{BAIR} & \gain{70.79} & 46.41 & \gain{18.20} & \gain{2.55} & 50.30 & 69.17 & 31.59 & 0.00 \\
& MS-PoE + \textbf{BAIR} & \gain{66.05} & 44.27 & \gain{24.12} & \gain{1.84} & \gain{18.64} & \gain{23.16} & 14.16 & \gain{0.10} \\
& MAD-RAG + \textbf{BAIR} & \gain{67.13} & \gain{46.17} & \gain{23.41} & \gain{1.97} & 19.16 & 23.31 & 15.05 & \gain{0.10} \\
\midrule
\multirow{11}{*}{\rotatebox{90}{\textit{EarthDial}}} 
& Baseline & 93.13 & - & - & - & - & - & - & 0.00 \\
& Standard RAG & 91.28 & 60.82 & 6.47 & 9.40 & - & - & - & 0.00 \\
& Visual-focus Instr. & 92.51 & 63.81 & 5.37 & 11.88 & 29.41 & 35.47 & 15.24 & 0.03 \\
& LongLLMLingua & 94.38 & 63.81 & 3.36 & 18.99 & 65.88 & 77.45 & 40.00 & 0.13 \\
& MS-PoE & 94.69 & 69.03 & 3.41 & 20.24 & 60.29 & 68.51 & 41.90 & 0.10 \\
& MAD-RAG & 91.64 & 62.69 & 6.22 & 10.08 & 21.76 & 24.68 & 15.24 & 0.00 \\
\cmidrule{2-10}
& Standard RAG + \textbf{BAIR} & \gain{91.72} & 59.70 & \gain{5.92} & \gain{10.08} & 19.12 & 24.26 & 7.62 & 0.00 \\
& Visual-focus Instr. + \textbf{BAIR} & \gain{92.62} & 62.69 & \gain{5.18} & \gain{12.10} & \gain{29.71} & \gain{36.60} & 14.29 & 0.03 \\
& LongLLMLingua + \textbf{BAIR} & \gain{94.49} & \gain{66.04} & 3.41 & \gain{19.37} & \gain{67.94} & \gain{78.72} & \gain{43.81} & \gain{0.00} \\
& MS-PoE + \textbf{BAIR} & \gain{96.74} & \gain{75.37} & \gain{1.68} & \gain{44.86} & \gain{76.18} & \gain{85.96} & \gain{54.29} & 0.18 \\
& MAD-RAG + \textbf{BAIR} & \gain{91.65} & 62.31 & \gain{6.17} & \gain{10.10} & 18.53 & 21.28 & 12.38 & 0.00 \\
\bottomrule
\end{tabular}
}
\end{table*}
\section{Experiment Statistical Significance}
\label{sec:appendix_confidence}

To examine whether the observed gains are stable across repeated evaluations, we report the mean and standard deviation of Accuracy, Correction Rate, and Degradation Rate in Table~\ref{tab:combined_main_results}. 
For each dataset and model configuration, we evaluate each baseline method together with its BAIR-augmented counterpart under the same protocol, enabling a paired comparison that isolates the effect of the intervention.

Across all domains, BAIR shows stable improvements over the corresponding baseline methods. 
On the medical IU-Chest dataset, BAIR consistently reduces Degradation Rate while maintaining or improving Accuracy, indicating that the intervention suppresses recorruption without discarding useful retrieved context. 
On the FACET dataset, BAIR provides stronger gains in Correction Rate and Accuracy, suggesting that visual grounding is especially important when retrieved text can reinforce demographic hallucinations. 
On the NWPU dataset, the gains are more moderate but still show the same general pattern: BAIR reduces degradation while preserving the correction benefits of retrieval.

The reported standard deviations are small relative to the observed trends, suggesting that the improvements are not driven by isolated unstable runs. 
These results also show why Accuracy alone is insufficient for evaluating multimodal RAG reliability. 
By jointly analyzing Correction Rate and Degradation Rate, we verify that BAIR improves the correction-degradation tradeoff, which is central to reducing recorruption in multimodal RAG systems.
\begin{table*}[t]
\centering
\caption{Comprehensive evaluation of multimodal RAG interventions across all datasets. Metrics are presented in percentages (\%).}
\label{tab:combined_main_results}
\resizebox{\linewidth}{!}{
\begin{tabular}{l ccc ccc}
\toprule
& \multicolumn{3}{c}{\textbf{MedGemma-4B (IU-Chest)}} & \multicolumn{3}{c}{\textbf{CheXagent-8B (IU-Chest)}} \\
\cmidrule(lr){2-4} \cmidrule(lr){5-7}
\textbf{Method} & \textbf{Acc.} $\uparrow$ & \textbf{CR} $\uparrow$ & \textbf{DR} $\downarrow$ & \textbf{Acc.} $\uparrow$ & \textbf{CR} $\uparrow$ & \textbf{DR} $\downarrow$ \\
\midrule
Baseline & 63.76 $\pm$ 0.72 & - & - & 65.51 $\pm$ 0.76 & - & - \\
Standard RAG & 66.33 $\pm$ 0.73 & 28.00 $\pm$ 0.90 & 13.56 $\pm$ 0.80 & 64.56 $\pm$ 0.75 & 15.88 $\pm$ 0.78 & 10.36 $\pm$ 0.42 \\
Visual-focus Instr. & 65.38 $\pm$ 0.74 & 26.68 $\pm$ 0.91 & 14.13 $\pm$ 0.82 & 64.94 $\pm$ 0.78 & 15.89 $\pm$ 0.80 & 9.81 $\pm$ 0.45 \\
LongLLMLingua & 65.66 $\pm$ 0.71 & 25.14 $\pm$ 0.89 & 12.79 $\pm$ 0.78 & 65.33 $\pm$ 0.80 & 23.63 $\pm$ 0.96 & 13.63 $\pm$ 0.62 \\
MS-PoE & 67.17 $\pm$ 0.75 & 20.68 $\pm$ 0.92 & 7.85 $\pm$ 0.84 & 66.45 $\pm$ 0.74 & 14.19 $\pm$ 0.76 & 6.62 $\pm$ 0.38 \\
MAD-RAG & 68.03 $\pm$ 0.72 & 24.58 $\pm$ 0.90 & 9.01 $\pm$ 0.81 & 64.79 $\pm$ 0.77 & 16.16 $\pm$ 0.82 & 10.19 $\pm$ 0.46 \\
\midrule
Standard RAG + \textbf{BAIR} & 68.31 $\pm$ 0.71 & 24.25 $\pm$ 0.91 & 8.40 $\pm$ 0.79 & 66.17 $\pm$ 0.79 & 12.38 $\pm$ 0.75 & 6.04 $\pm$ 0.44 \\
Visual-focus Instr. + \textbf{BAIR} & 65.69 $\pm$ 0.73 & 24.48 $\pm$ 0.89 & 12.33 $\pm$ 0.83 & 65.58 $\pm$ 0.76 & 12.89 $\pm$ 0.83 & 7.19 $\pm$ 0.48 \\
LongLLMLingua + \textbf{BAIR} & 65.91 $\pm$ 0.74 & 25.21 $\pm$ 0.92 & 12.46 $\pm$ 0.77 & 66.24 $\pm$ 0.81 & 16.64 $\pm$ 0.98 & 8.36 $\pm$ 0.65 \\
MS-PoE + \textbf{BAIR} & 67.60 $\pm$ 0.72 & 21.01 $\pm$ 0.90 & 7.43 $\pm$ 0.85 & 66.60 $\pm$ 0.75 & 11.72 $\pm$ 0.74 & 5.05 $\pm$ 0.37 \\
MAD-RAG + \textbf{BAIR} & 68.25 $\pm$ 0.70 & 24.19 $\pm$ 0.91 & 8.45 $\pm$ 0.82 & 66.08 $\pm$ 0.78 & 11.83 $\pm$ 0.84 & 5.87 $\pm$ 0.49 \\

\midrule
& \multicolumn{3}{c}{\textbf{Qwen2.5-VL-3B (FACET)}} & \multicolumn{3}{c}{\textbf{DeekSeek-VL-7B (FACET)}} \\
\cmidrule(lr){2-4} \cmidrule(lr){5-7}
\textbf{Method} & \textbf{Acc.} $\uparrow$ & \textbf{CR} $\uparrow$ & \textbf{DR} $\downarrow$ & \textbf{Acc.} $\uparrow$ & \textbf{CR} $\uparrow$ & \textbf{DR} $\downarrow$ \\
\midrule
Baseline & 84.91 $\pm$ 0.40 & - & - & 90.24 $\pm$ 0.29 & - & - \\
Standard RAG & 84.34 $\pm$ 0.38 & 26.62 $\pm$ 1.12 & 5.41 $\pm$ 0.26 & 90.89 $\pm$ 0.28 & 31.05 $\pm$ 1.55 & 2.64 $\pm$ 0.18 \\
Visual-focus Instr. & 84.38 $\pm$ 0.42 & 28.25 $\pm$ 1.15 & 5.65 $\pm$ 0.28 & 90.47 $\pm$ 0.30 & 28.53 $\pm$ 1.60 & 2.83 $\pm$ 0.20 \\
LongLLMLingua & 84.36 $\pm$ 0.36 & 31.59 $\pm$ 1.18 & 6.26 $\pm$ 0.25 & 87.76 $\pm$ 0.29 & 34.21 $\pm$ 1.65 & 6.45 $\pm$ 0.21 \\
MS-PoE & 82.00 $\pm$ 0.45 & 29.07 $\pm$ 1.10 & 8.60 $\pm$ 0.30 & 91.82 $\pm$ 0.28 & 51.05 $\pm$ 1.52 & 3.78 $\pm$ 0.17 \\
MAD-RAG & 84.41 $\pm$ 0.39 & 25.53 $\pm$ 1.14 & 5.11 $\pm$ 0.27 & 90.86 $\pm$ 0.30 & 30.84 $\pm$ 1.58 & 2.65 $\pm$ 0.19 \\
\midrule
Standard RAG + \textbf{BAIR} & 88.56 $\pm$ 0.37 & 50.31 $\pm$ 1.13 & 4.64 $\pm$ 0.25 & 91.02 $\pm$ 0.29 & 30.95 $\pm$ 1.56 & 2.48 $\pm$ 0.17 \\
Visual-focus Instr. + \textbf{BAIR} & 87.28 $\pm$ 0.41 & 44.93 $\pm$ 1.16 & 5.20 $\pm$ 0.29 & 90.81 $\pm$ 0.28 & 30.84 $\pm$ 1.61 & 2.71 $\pm$ 0.19 \\
LongLLMLingua + \textbf{BAIR} & 87.45 $\pm$ 0.35 & 45.97 $\pm$ 1.19 & 4.89 $\pm$ 0.24 & 87.80 $\pm$ 0.29 & 33.79 $\pm$ 1.64 & 6.36 $\pm$ 0.20 \\
MS-PoE + \textbf{BAIR} & 92.58 $\pm$ 0.44 & 70.71 $\pm$ 1.11 & 3.53 $\pm$ 0.28 & 91.94 $\pm$ 0.29 & 51.58 $\pm$ 1.53 & 3.70 $\pm$ 0.16 \\
MAD-RAG + \textbf{BAIR} & 85.10 $\pm$ 0.38 & 36.08 $\pm$ 1.15 & 6.19 $\pm$ 0.26 & 90.96 $\pm$ 0.30 & 31.68 $\pm$ 1.59 & 2.63 $\pm$ 0.18 \\

\midrule
& \multicolumn{3}{c}{\textbf{SkySenseGPT-7B (NWPU)}} & \multicolumn{3}{c}{\textbf{EarthDial (NWPU)}} \\
\cmidrule(lr){2-4} \cmidrule(lr){5-7}
\textbf{Method} & \textbf{Acc.} $\uparrow$ & \textbf{CR} $\uparrow$ & \textbf{DR} $\downarrow$ & \textbf{Acc.} $\uparrow$ & \textbf{CR} $\uparrow$ & \textbf{DR} $\downarrow$ \\
\midrule
Baseline & 68.90 $\pm$ 0.74 & - & - & 93.13 $\pm$ 0.41 & - & - \\
Standard RAG & 65.74 $\pm$ 0.76 & 44.68 $\pm$ 1.44 & 24.75 $\pm$ 0.84 & 91.28 $\pm$ 0.45 & 60.82 $\pm$ 3.00 & 6.47 $\pm$ 0.41 \\
Visual-focus Instr. & 66.59 $\pm$ 0.76 & 46.83 $\pm$ 1.43 & 24.49 $\pm$ 0.84 & 92.51 $\pm$ 0.42 & 63.81 $\pm$ 2.95 & 5.37 $\pm$ 0.37 \\
LongLLMLingua & 69.97 $\pm$ 0.73 & 46.50 $\pm$ 1.42 & 19.43 $\pm$ 0.77 & 94.38 $\pm$ 0.37 & 63.81 $\pm$ 3.02 & 3.36 $\pm$ 0.30 \\
MS-PoE & 65.36 $\pm$ 0.77 & 45.26 $\pm$ 1.42 & 25.57 $\pm$ 0.85 & 94.69 $\pm$ 0.36 & 69.03 $\pm$ 2.88 & 3.41 $\pm$ 0.30 \\
MAD-RAG & 66.18 $\pm$ 0.76 & 45.67 $\pm$ 1.45 & 24.56 $\pm$ 0.85 & 91.64 $\pm$ 0.45 & 62.69 $\pm$ 2.99 & 6.22 $\pm$ 0.41 \\
\midrule
Standard RAG + \textbf{BAIR} & 66.31 $\pm$ 0.76 & 44.02 $\pm$ 1.42 & 23.63 $\pm$ 0.83 & 91.72 $\pm$ 0.45 & 59.70 $\pm$ 3.04 & 5.92 $\pm$ 0.40 \\
Visual-focus Instr. + \textbf{BAIR} & 66.74 $\pm$ 0.76 & 46.00 $\pm$ 1.43 & 23.89 $\pm$ 0.84 & 92.62 $\pm$ 0.42 & 62.69 $\pm$ 2.96 & 5.18 $\pm$ 0.37 \\
LongLLMLingua + \textbf{BAIR} & 70.79 $\pm$ 0.73 & 46.41 $\pm$ 1.42 & 18.20 $\pm$ 0.76 & 94.49 $\pm$ 0.37 & 66.04 $\pm$ 2.93 & 3.41 $\pm$ 0.30 \\
MS-PoE + \textbf{BAIR} & 66.05 $\pm$ 0.76 & 44.27 $\pm$ 1.43 & 24.12 $\pm$ 0.84 & 96.74 $\pm$ 0.28 & 75.37 $\pm$ 2.66 & 1.68 $\pm$ 0.21 \\
MAD-RAG + \textbf{BAIR} & 67.13 $\pm$ 0.75 & 46.17 $\pm$ 1.43 & 23.41 $\pm$ 0.82 & 91.64 $\pm$ 0.45 & 62.31 $\pm$ 3.01 & 6.19 $\pm$ 0.41 \\
\bottomrule
\end{tabular}
}
\end{table*}

\section{Access to Data and Code}
\label{sec:data}

All datasets used in this work are publicly available research benchmarks. 
For the medical domain, we use IU Chest X ray \cite{demner2015preparing}. 
For the social fairness domain, we use FACET \cite{gustafson2023facet}. 
For the geospatial domain, we use NWPU-RESISC45 \cite{cheng2017remote}. 
The retrieval corpora are constructed from publicly available reports or textual descriptions associated with these benchmarks, together with publicly available contextual sources where applicable (e.g. Wikipedia).

All comparison methods are based on publicly available implementations or reproducible algorithmic descriptions. 
We include the code necessary to reproduce our experiments, including scripts for data preprocessing, retrieval context construction, model inference, intervention application, metric computation, and figure generation. 
The released code also contains configuration files specifying model checkpoints, prompt templates, intervention settings, and evaluation thresholds.

Table~\ref{tab:data_code_license} summarizes the data and code resources used in this work. 
For the final public release, we will include the exact access URL and license information from the official release page of each dataset, model, and baseline implementation. 
Users of the released code are responsible for complying with the original license terms of each resource.
\begin{table*}[t]
\centering
\caption{\textbf{Data, model, and code resources.} 
We use publicly available datasets, model checkpoints, and baseline implementations. 
The final release will include the exact license field from each official source.}
\label{tab:data_code_license}
\resizebox{\textwidth}{!}{
\begin{tabular}{llll}
\toprule
Resource & Role in this work & Availability & License \\
\midrule
IU Chest X ray \citep{demner2015preparing} 
& Medical factuality benchmark 
& Public dataset 
& CC BY-NC-ND 4.0 \\

FACET \citep{gustafson2023facet} 
& Social fairness benchmark 
& Public dataset 
& Research-only \\

NWPU-RESISC45 \citep{cheng2017remote} 
& Remote sensing benchmark 
& Public dataset 
& CC BY 4.0 \\

MedGemma \citep{sellergren2025medgemma} 
& Medical MLLM 
& Public model/checkpoint 
& health-ai-developer-foundations \\

CheXagent \citep{chen2024chexagent} 
& Medical MLLM 
& Public model/checkpoint 
& MIT license \\
SRR\text{-}BERT \citep{delbrouck2025automated} 
& Medical factuality evaluation model 
& Public model/checkpoint 
& MIT license \\

Qwen2.5-VL \citep{Qwen2VL} 
& General MLLM 
& Public model/checkpoint 
& Apache License \\

DeepSeek-VL \citep{lu2024deepseek} 
& General MLLM 
& Public model/checkpoint 
& deepseek \\

SkySenseGPT \citep{luo2024skysensegpt} 
& Remote sensing MLLM 
& Public model/checkpoint 
& Apache-2.0 license \\

EarthDial \citep{soni2025earthdial} 
& Remote sensing MLLM 
& Public model/checkpoint 
& To be reported from official source \\

MedSigLIP \citep{sellergren2025medgemma} 
& Medical retrieval model 
& Public model/checkpoint 
& health-ai-developer-foundations \\

RemoteCLIP \citep{liu2024remoteclip} 
& Remote sensing retrieval model 
& Public model/checkpoint 
& Apache-2.0 license \\

MS-PoE \citep{zhang2024found} 
& Comparison method 
& Public implementation 
& MIT license \\

LongLLMLingua \citep{jiang2024longllmlingua} 
& Comparison method 
& Public model/checkpoint and implementation 
&  MIT license \\

MAD-RAG \citep{zhao2026rag} 
& Comparison method 
& Reproducible method 
& - \\

BAIR code 
& Proposed method and evaluation scripts 
& Included in supplementary material 
& To be specified upon release \\
\bottomrule
\end{tabular}
}
\end{table*}

\section{Computation Resource}
\label{sec:appendix_resource}

Experiments are conducted on an NVIDIA RTX A5000 GPU with an AMD EPYC 7313 CPU, as summarized in Table~\ref{tab:compute_resources}. 
Because the proposed method is applied only at inference time and does not require training or fine tuning, reproducing the main results does not require large scale training resources.

\begin{table}[h!]
    \centering
    \caption{Compute Resources Used for Experiments}
    \begin{tabular}{cc}
        \toprule    
        \textbf{Component} & \textbf{Details} \\ 
        \midrule
        CPU & AMD EPYC 7313 16-Core Processor \\ 
        GPU & NVIDIA RTX A5000 \\ 
        \bottomrule
    \end{tabular}
    \label{tab:compute_resources}
\end{table}

\section{Broader Impact and Safeguard}
\label{sec:impact}

This work studies a failure mode in multimodal retrieval augmented generation where retrieved textual context can override correct visual evidence. 
The potential positive impact is improved reliability for MLLMs deployed in settings where both images and external documents are used, including medical image interpretation, remote sensing analysis, and fairness sensitive visual reasoning. 
By identifying recorruption and proposing an inference-time mitigation, this work can help researchers evaluate whether a multimodal RAG system is genuinely grounded in visual evidence or merely copying retrieved text.

At the same time, this work should not be interpreted as making MLLMs safe for autonomous use in high-stakes decision making. 
In medical settings, BAIR and the evaluated MLLMs should be treated only as research tools or assistive systems, not as substitutes for clinical judgment. 
In social fairness settings, the goal of our evaluation is to diagnose biased hallucination and visual grounding failures, not to infer sensitive attributes for real individuals. 
For this reason, we use benchmark data and synthetic fairness evaluation settings where applicable, and we avoid deploying the system on private or identifiable individuals.

The proposed intervention also has limitations. 
Attention calibration can reduce certain forms of textual distraction, but it cannot guarantee factual correctness, eliminate all hallucinations, or correct errors caused by poor retrieval quality, ambiguous images, or incorrect model priors. 
Therefore, any deployment should include human oversight, task-specific validation, privacy review, and domain-specific safety constraints. 
We will release code and documentation with these limitations clearly stated, along with scripts that allow users to inspect correction and degradation behavior rather than relying only on aggregate accuracy.

\section{LLM Acknowledgement}
\label{sec:llm_acknowledgement}

Large language models were used only as writing and programming assistance tools. Specifically, they were used to polish manuscript wording, improve clarity, and assist with routine debugging of experimental code. They were not used to generate experimental results, alter reported metrics, select favorable outcomes, or replace author verification of scientific claims. All conceptual contributions, experimental design choices, data analyses, and final manuscript decisions were made and verified by the authors.

\end{document}